%% file: root.tex
\documentclass[journal]{IEEEtran}

\usepackage{amsmath,amsfonts}
\usepackage{algorithmic}
\usepackage{array}
\usepackage[caption=false,font=normalsize,labelfont=sf,textfont=sf]{subfig}
\usepackage{textcomp}
\usepackage{stfloats}
\usepackage{url}
\usepackage{verbatim}
\usepackage{graphicx}
\usepackage{cite}
\usepackage{xcolor} 
\usepackage{booktabs} 
\usepackage{multirow}
\usepackage{caption} 
\usepackage{cuted}
\usepackage{enumitem}
\usepackage{gensymb}
\usepackage{siunitx}
\usepackage{eso-pic}

\usepackage[nohyperlinks, printonlyused, nolist]{acronym}
\input{chapters/acronyms}

\begin{document}
\AddToShipoutPictureFG*{\AtPageUpperLeft{\put(\LenToUnit{0.5\paperwidth},\LenToUnit{-0.22in}){%
  \makebox[0pt]{\parbox[t]{0.86\paperwidth}{\centering\scriptsize
  This article has been accepted for publication in IEEE Transactions on Robotics.
  \copyright~2026 IEEE. Personal use of this material is permitted. Permission from IEEE must be obtained for all other uses,
  in any current or future media, including reprinting/republishing this material for advertising or promotional purposes,
  creating new collective works, for resale or redistribution to servers or lists, or reuse of any copyrighted component
  of this work in other works.}}}}}
\bstctlcite{IEEEexample:BSTcontrol}
\title{DeFM: Learning Foundation Representations from Depth for Robotics}

\newif\ifanonymous
\anonymousfalse 

\DeclareRobustCommand{\anontext}[2]{%
  \ifanonymous
    \textcolor{orange}{[#2]}%
  \else
    #1%
  \fi
}

\DeclareRobustCommand{\anon}[1]{%
  \ifanonymous
  \else
    #1%
  \fi
}

\newcommand{\finaltodo}[1]{\textcolor{red}{[TODO: #1]}} 
\newcommand{\etal}{\textit{et al}. }
\newcommand{\ie}{\textit{i}.\textit{e}., }
\newcommand{\eg}{\textit{e}.\textit{g}. }

\author{
\anontext{
Manthan Patel$^{1*}$, Jonas Frey$^{2}$, Mayank Mittal$^{1, 3}$, Fan Yang$^{1}$, Alexander
Hansson$^{1}$, \\ Amir Bar$^{4}$, Cesar Cadena$^{1}$, and Marco Hutter$^{1}$
\thanks{$^1$ Robotic Systems Lab (RSL), ETH Zurich, Switzerland}
\thanks{$^2$ UC Berkeley, Berkeley, USA}
\thanks{$^3$ NVIDIA, Zurich, Switzerland}
\thanks{$^4$ Imperial College London, UK}
\thanks{$^*$ Corresponding: \texttt{patelm@ethz.ch}}
}{anonymous authors}
}



\maketitle

\begin{strip}
    \vspace{-3\baselineskip}
    \centering
    \includegraphics[width=\textwidth]{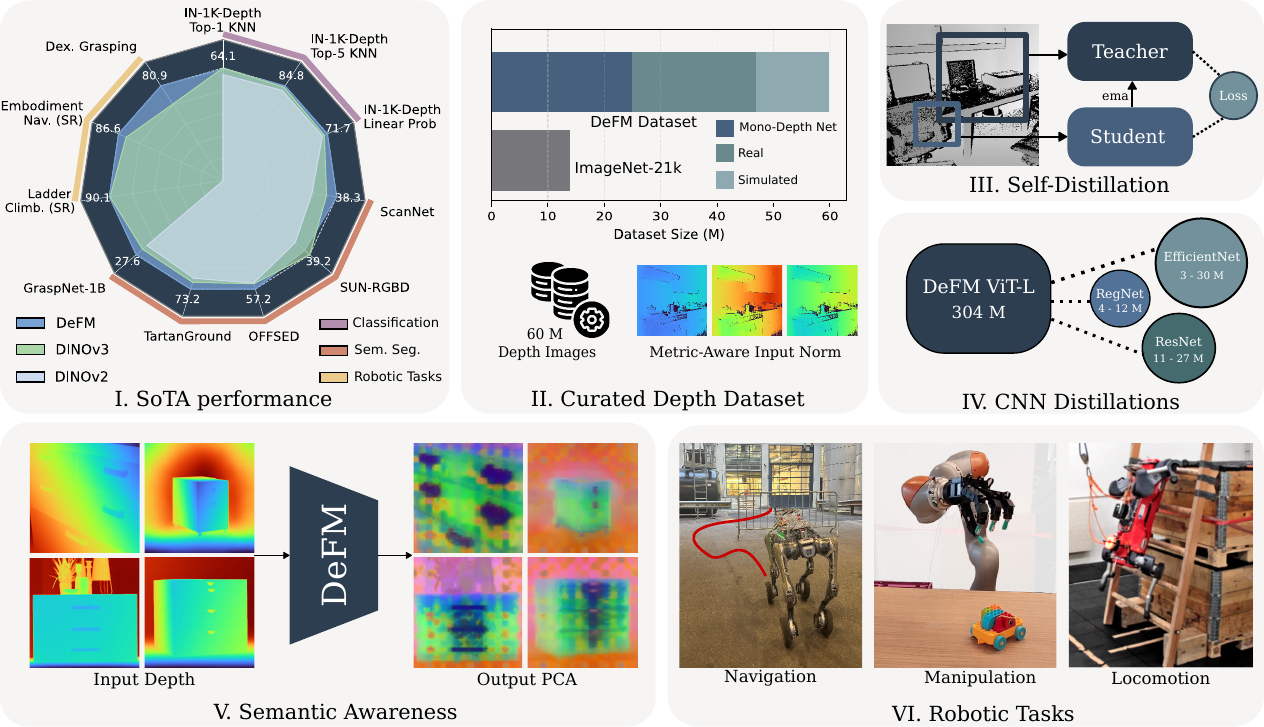}
    \captionof{figure}{We present DeFM, a foundation model for depth images.
    Pretrained using DINOv2 style self-distillation method (III) on a curated depth dataset of
    \SI{60}{M} images (II), DeFM features achieve state-of-the-art results across several
    classification and semantic segmentation benchmarks (linear probing) (I).
    Features obtained from DeFM reveal semantic awareness upon performing PCA despite depth lacking
    texture and color (V).
    We distill our largest DeFM model into several efficient CNN networks (IV) to be used for
    various downstream robotic Reinforcement Learning tasks, including navigation, manipulation, and
    locomotion (VI).
    In (I), the scales are linearly scaled to the top performer, and zero performance implies it was
    not tested.}
    \label{fig:highlight}
    \vspace{0.5em}
\end{strip}

\thispagestyle{empty}
\pagestyle{empty}
\begin{abstract}
\input{chapters/0_abstract}

\end{abstract}


\section{Introduction}
\label{sec:Introduction}
\input{chapters/1_introduction}

\section{Related Work}
\label{sec:Related_work}
\input{chapters/2_related_work}

\section{Preliminaries}
\label{sec:Prelimnary}
\input{chapters/3_prelimnary}

\section{Depth Foundation Model}
\label{sec:Method}
\input{chapters/4_method}

\section{Experiments}
\label{sec:Experiments}
\input{chapters/5_experiments}

\section{Robotic Experiments}
\label{sec:Robotic_experiments}
\input{chapters/6_robotic_experiments}


\section{Limitations and Future Work}
\label{sec:Limitations}
\input{chapters/7_limitations}

\section{Conclusion}
\label{sec:Conclusion}
\input{chapters/8_conclusion}

\appendices
\section{Implementation Details}
\label{app:impl}
\input{chapters/9_appendix}

\anon{\input{chapters/acknowledgements}}

\bibliographystyle{IEEEtran}
\bibliography{bibliography}

\input{chapters/10_biography}

\end{document}

%% file: chapters/acronyms.tex
\begin{acronym}
\acro{AI}[AI]{Artificial Intelligence}
\acro{LLM}[LLM]{Large Language Model}
\acro{VLA}{Vision Language Action}
\acro{VLM}{Vision Language Model}
\acro{RT}{Real-Time}
\acro{GT}{Ground Truth}
\acro{GNSS}{Global Navigation Satellite System}
\acro{RTK}{Real-time Kinematic Positioning}
\acro{LiDAR}{Light Detection and Ranging}
\acro{IMU}{Inertial Measurement Unit}
\acro{RPE}{Relative Position Error}
\acro{RRE}{Relative Rotation Error}
\acro{APE}{Absolute Position Error}
\acro{ARE}{Relative Rotation Error}
\acro{RL}{Reinforcement Learning}
\acro{HF}{Holistic Fusion}
\acro{MAP}{Maximum a Posteriori}
\acro{SLAM}{Simultaneous Localization and Mapping}
\acro{SSL}{Self-Supervised Learning}
\acro{VFM}{Vision Foundation Model}
\acro{BiFPN}{Bi-directional Feature Pyramid Network}
\acro{PCA}{Principal Component Analysis}
\acro{ViT}{Vision Transformer}
\end{acronym}

%% file: chapters/0_abstract.tex
Depth sensors are widely deployed across robotic platforms, and advances in fast, high-fidelity
depth simulation have enabled robotic policies trained on depth observations to achieve robust
sim-to-real transfer for a wide range of tasks.
Despite this, representation learning for depth modality remains underexplored compared to RGB,
where large-scale foundation models now define the state of the art.
To address this gap, we present DeFM, a self-supervised foundation model trained entirely on
depth images for robotic applications.
Using a DINO-style self-distillation objective on a curated dataset of 60M depth images, DeFM learns
geometric and semantic representations that generalize to diverse environments, tasks, and sensors.
To retain metric awareness across multiple scales, we introduce a novel input normalization
strategy.
We further distill DeFM into compact models suitable for resource-constrained robotic systems. 
When evaluated on depth-based classification, segmentation, navigation, locomotion, and
manipulation, DeFM achieves state-of-the-art performance across several benchmarks and demonstrates
strong generalization from simulation to real-world environments.
We release all our pretrained models, which can be adopted off-the-shelf for depth-based robot
learning.
\textit{\anon{Webpage: \url{https://de-fm.github.io/}}}

%% file: chapters/1_introduction.tex
Foundation models pretrained on large-scale RGB datasets have transformed visual representation
learning in recent years.
Methods such as DINO~\cite{caron2021emerging, oquab2023dinov2, simeoni2025dinov3} and
CLIP~\cite{radford2021learning} have become the de facto standard for vision tasks, demonstrating
remarkable zero-shot transfer to classification, detection, segmentation, and retrieval.
Their success has extended into robotics, where pretrained RGB encoders serve as the backbone for
various tasks such as manipulation, navigation, and scene
understanding~\cite{kim24openvla, intelligence2025pi_, zhang2024uni}.
Specialized \acp{VFM} tailored to
robotics~\cite{shang2024theia, Nair2022R3MAU, gavryushin2025maple, xiao2022masked} further
highlight the central role of large-scale self-supervised representation learning in enabling
embodied agents.

However, RGB-only observations are challenging to use for robotics, particularly in the context of
sim-to-real transfer, which is a central paradigm in robot learning.
This difficulty arises because simulated RGB rarely captures real-world visual complexity.
In contrast, depth is invariant to lighting, texture, and color variation, establishing it as highly
effective for generalization and sim-to-real transfer~\cite{muratore2022robot}.
Depth sensors are now ubiquitous across robotic platforms, including manipulators, mobile robots,
drones, and AR/VR devices, and have been leveraged for
locomotion~\cite{miki2022learning, rudin2022learning, agarwal2023legged},
navigation~\cite{Wijmans2019DDPPOLN, lee2024learning, yang2025improving}, and
manipulation~\cite{lum2024dextrahg, singh2025end, zhang2025robustdexgrasp}.
This widespread adoption highlights depth as an important modality for scalable representation
learning in robotics, especially when robustness and generalization are required.

Despite the unique advantages of depth modality, to the best of our knowledge, currently no general-purpose depth encoders exist.
Current approaches either re-purpose RGB-pretrained encoders for depth by applying colormaps or
channel stacking~\cite{hu2019acnet}, which introduces distribution mismatch and degrades geometric
fidelity, or they train task-specific depth encoders from
scratch~\cite{agarwal2023legged, lum2024dextrahg}, leading to poor generalization across tasks,
environments, and sensors.
Self-supervised learning on depth remains underexplored, despite the fact that depth inherently
captures critical cues for robotics, including metric geometry (e.g., object size and distance),
free space (e.g., occupancy), and physical affordances, which are the action possibilities an object
offers a robot, such as grasping.
Given the availability and advantages of depth data, we raise the natural question: \par 
\vspace{2mm}

\emph{Can we build a universal pretrained depth encoder for robotics?}
\par
\vspace{2mm}
The simple answer is yes.
In this work, we present DeFM, a self-supervised foundation model trained exclusively on
depth images that provides transferable features for embodied agents.
As a first qualitative result, we show that the feature space learned by DeFM is fundamentally
sound: embeddings from the depth encoder reveal emergent semantic clustering when visualized via PCA
(Fig.~\ref{fig:pca_cups}), mirroring the structure observed in effective RGB representations.
This suggests that, even without texture or color, semantically meaningful structure
can be extracted from depth, beyond pure geometry.
\par

DeFM builds on the DINOv2 self-distillation framework and adapts it to the depth domain at scale.
Our contributions are therefore depth-specific: a pretraining dataset combining real, synthetic, and
Monocular Depth Estimation (MDE)-derived depth; an input normalization designed for spatial scales
ranging from fine-grained manipulation to long-range navigation; and the distillation of our largest
model into compact CNN and ViT encoders for resource-constrained deployment.
\par

Most critically, we demonstrate that DeFM can be applied \emph{frozen} across a broad spectrum of
robotics tasks.
Evaluated on depth-based classification, segmentation, locomotion, navigation, and manipulation,
DeFM outperforms RGB-pretrained encoders repurposed for depth (Fig.~\ref{fig:highlight}) and
recent point cloud foundation models in most settings, with robust sim-to-real transfer.
In summary, we make the following key contributions:

\begin{itemize}
    \item We present DeFM, a large-scale self-supervised foundation model for
        depth image inputs with a focus on robotic tasks.
    \item We curate a dataset of \SI{60}{M} depth images and release distilled models ranging from
        \SI{3}{M} to \SI{30}{M} parameters, covering both CNNs and ViTs, to enable wide adoption in
        robotics research.
    \item We introduce a novel input normalization strategy to retain metric awareness across
        multiple scales.
\end{itemize}

%% file: chapters/2_related_work.tex
\subsection{Vision Foundation Models}
\label{sec:lit_review_vfm}

Early visual representation learning relied on supervised classification, but such features often
transfer sub-optimally to dense spatial prediction
tasks~\cite{dosovitskiy2020image, balestriero2023cookbook}.
This limitation motivated a paradigm shift toward \ac{SSL}, which trains large \acp{VFM} on raw data
without human annotations to yield highly transferable, domain-flexible representations.

\subsubsection{Self-Supervised Pretraining Paradigms}
Early contrastive frameworks~\cite{chen2020simple, he2020momentum} learned representations by
maximizing the similarity between different augmented views of the same image using large negative
sets.
To avoid negative pairs, self-distillation methods instead optimize a student network to predict the
latent representations of a momentum-updated teacher~\cite{grill2020bootstrap, caron2021emerging}.
Concurrently, MAE~\cite{he2022masked} reconstructs masked image patches to force the encoder to
capture global context.
Modern state-of-the-art frameworks unify these concepts: methods such as iBOT~\cite{zhou2021ibot}
and DINOv2~\cite{oquab2023dinov2} combine self-distillation with masked prediction over massive
curated datasets, yielding highly transferable dense features for diverse downstream tasks.
More recently, in DINOv3~\cite{simeoni2025dinov3}, the authors further scale up the training dataset
to an order of magnitude higher (1.7B images) and introduce gram anchoring to prevent degradation of
dense feature maps during a long training schedule.
Alternatively, architectures like I-JEPA~\cite{assran2023self,bar2024stochastic} adopt a
non-generative masked prediction objective that predicts high-level latent representations rather
than pixels, leading to semantically rich and stable foundational features. 

\subsubsection{Vision Language and Geometric Foundation Models}

Vision-Language Models (VLMs), such as CLIP~\cite{radford2021learning} and its
variants~\cite{zhai2023sigmoid, jia2021scaling}, introduce a powerful form of self-supervision by
learning to align visual representations with corresponding natural language captions using a
contrastive loss on massive web-scraped datasets.
This objective grants the model exceptional zero-shot transfer capabilities and strong semantic
grounding, allowing it to solve classification and retrieval tasks using only text prompts.
However, while VLMs are semantically rich, their representations are often less spatially and
geometrically detailed than those produced by dense visual SSL models like
DINOv2~\cite{el2024probing}.

More recently, geometric foundation models such as DUSt3R~\cite{wang2024dust3r},
VGGT~\cite{wang2025vggt}, and DA3~\cite{lin2025depth} have emerged, demonstrating powerful
multi-view 3D reconstruction and depth estimation directly from unconstrained RGB inputs.
Alternatively, a complementary trend is Agglomerative Vision Models, such as RADIO (Reduce All
Domains Into One) \cite{Ranzinger_2024_CVPR, heinrich2025radiov25improvedbaselinesagglomerative},
which train a single student model by distilling the knowledge from multiple pre-trained,
heterogeneous \acp{VFM} (e.g., CLIP, DINOv2, SAM~\cite{kirillov2023segment}).
This multi-teacher distillation approach efficiently consolidates the diverse strengths, such as
zero-shot text grounding and dense correspondence, into a unified encoder, creating a single,
versatile backbone for downstream tasks.
However, all these existing \acp{VFM} predominantly rely on RGB inputs, leaving a critical gap for a
VFM operating on depth images.
Our work addresses this by adapting DINOv2 self-supervision to learn dense spatial representations
for robotics tasks.

\subsection{RGB-D Representation Learning}

While standalone depth representation learning remains underexplored, prior works have explored
utilizing the complementary strengths of RGB and depth images by joint encoding to obtain better
representations.
Architectures like DFormer~\cite{yin2024dformer} and its extension DFormer++~\cite{yin2026dformer++}
introduced dedicated RGB-D pretraining frameworks that utilize image-depth pairs, explicitly
providing the backbone with the capacity to encode true 3D geometric relationships.
Other recent frameworks have optimized this fusion further with AsymFormer~\cite{du2024asymformer}
utilizing an asymmetrical cross-modal backbone for highly efficient, real-time RGB-D segmentation.
Beyond segmentation, RGB-D representation learning has been effectively decoupled for complex
spatiotemporal tasks such as motion recognition~\cite{zhou2022decoupling}, multi-modal object
detection~\cite{xu2017multi}, and grasping~\cite{qin2023rgb}.
More recently, Lingbot-Depth~\cite{tan2026masked} introduces Masked Depth Modeling to align the RGB
and Depth features, demonstrating generalization for several downstream tasks such as depth
completion, point tracking, and robot grasp pose generation.
However, because these methods inherently require paired RGB and depth inputs, developing universal
foundation models trained exclusively on standalone depth images remains unaddressed.

\subsection{Point Cloud Foundation Models}

As a direct 3D counterpart to 2D depth representation learning, recent advancements have introduced
powerful foundation models for point clouds.
Frameworks such as Masked Scene Contrast~\cite{wu2023masked}, Sonata~\cite{wu2025sonata}, and
Concerto~\cite{zhang2026concerto} have established strong baselines for 3D scene understanding
through masked modeling, contrastive pretraining, and 2D-3D cross-modal joint
embeddings~\cite{zhang2026concerto}.
Despite these successes, the pretraining corpora for these models remain heavily constrained to
specific domains, primarily indoor environments and autonomous driving scenes.
While recent efforts like Utonia~\cite{zhang2026utonia} attempt to unify diverse domains, including
object-centric point clouds and manipulation datasets, they still lack the broad, cross-domain
diversity required for generalized robotic tasks.
Furthermore, these architectures predominantly rely on variants of Point Transformer v3
(PTv3)~\cite{wu2024ptv3}, which introduces significant computational overhead and latency, making
them difficult to deploy in high-frequency, real-time robotic control loops.
Although emerging architectures like the Lite Point Transformer~\cite{yuelitept2026} aim to mitigate
these bottlenecks, the availability of truly lightweight, off-the-shelf 3D encoders tailored
specifically for robotics applications remains highly limited.

\subsection{Visual Representations for Robot Learning}

The features learned by early RGB VFMs are often insufficient for robotics, which requires grounded,
action-relevant features emphasizing geometry and physical affordances over general
semantics~\cite{majumdar2023we}.
This limitation has driven the development of pretrained visual representations specifically
tailored for embodied intelligence.

\subsubsection{Action-Centric Pretraining} 
A key approach involves pretraining encoders on large-scale action-centric video datasets to learn
representations with temporal and kinematic priors.
R3M~\cite{Nair2022R3MAU} performs pretraining on in-the-wild video using contrastive and
video-language objectives.
R3M's frozen encoder improves manipulation success rates by 10-20\% over scratch or CLIP baseline
models on a suite of tasks.
Similarly, VIP~\cite{ma2022vip} uses a value-implicit objective, leveraging the relationship between
visual observations and expected future rewards to learn representations that inherently capture
value and utility for control.
More recently MAPLE~\cite{gavryushin2025maple} exploits rich manipulation priors learned from
large-scale egocentric videos by predicting fine-grained hand-object contact points and hand poses
to enable efficient policy learning for complex downstream tasks.

\subsubsection{Generative and Predictive Modeling}

Another major direction adapts the Masked Image Modeling objective to the robotics domain.
MVP~\cite{xiao2022masked} trains \ac{ViT} encoders by reconstructing masked patches in egocentric
videos.
This technique was scaled to real-world scenarios \cite{radosavovic2023real} and extended in models
like Multi-view Masked World Models \cite{seo2023multi} to leverage multi-camera views, enabling
better 3D consistency.
More advanced geometric extensions, such as 3D-MVP \cite{qian20243d}, explicitly integrate geometric
information from multiple views during pretraining to improve 3D spatial reasoning for manipulation.
The emergence of 4D Visual Pre-training (FVP) \cite{hou20254d} further highlights the importance of
incorporating spatio-temporal dynamics into the pretraining objective. 

\subsubsection{Zero-Shot Transfer of VFMs for Robotics}

With the advancements of VFMs, they are now being increasingly used for various robotic tasks
without fine-tuning.
For instance, DINOBot~\cite{di2024dinobot} uses DINO features for semantic retrieval and pixel-level
alignment to enable one-shot imitation learning, significantly improving data efficiency.
Other methods integrate DINOv2 features into voxel representations for bimanual manipulation
\cite{yurchyk2025large} or use them for zero-shot planning within a World Model
framework~\cite{zhou2024dino}.
This suggests that the dense, robust feature space learned by DINOv2 is a strong prior for robotics.
Some works, such as Theia~\cite{shang2024theia}, propose distilling the knowledge from multiple
different VFMs into a single, smaller backbone showcasing improvements over a single VFM backbone
across diverse robotics tasks.

\subsection{Reinforcement Learning from Depth}

Depth sensing, in contrast to RGB, provides direct access to metric, three-dimensional geometry,
which is crucial for grounded robotic interaction and robust sim-to-real transfer
\cite{muratore2022robot, loquercio2021learning, yu2025depth}.
This property has led to extensive research focusing on training end-to-end RL policies directly
from depth for various manipulation, locomotion, and navigation tasks.

\subsubsection{Locomotion and Navigation}

In legged and mobile robotics, accurate knowledge of terrain geometry is crucial for robust
locomotion and navigation.
Earlier works addressed perceptive locomotion by converting depth sensor data into explicit
geometric structures, such as height maps or elevation maps
\cite{miki2022learning, rudin2022learning, lee2024learning, zhang2024resilient}.
While these explicit maps are straightforward to render in simulation, they often suffer from
significant noise, require highly accurate odometry, and are challenging to reliably obtain in
real-world scenarios due to the limited Field of View (FoV) of ego-centric depth cameras, frequently
leading to a sim-to-real gap~\cite{he2025attention}.
To address these limitations, a major shift involved learning end-to-end policies directly from
egocentric depth observations via RL.
Agarwal~\etal~\cite{agarwal2023legged} were among the first ones to demonstrate an end-to-end
locomotion system capable of traversing challenging terrains (stairs, curbs, stepping stones) using
a quadruped robot equipped with only a single front-facing depth camera.
Subsequent methods have built on this foundation, utilizing egocentric depth for more complex
behaviors such as parkour~\cite{rudin2025parkour, luo2024pie} and learning local obstacle
avoidance~\cite{kareer2022vinl}.
More recently, this paradigm has been extended to humanoid locomotion, where the robustness
requirements are even more stringent.
Approaches~\cite{zhuang2024humanoid, sun2025dpl} train depth-only RL policies and focus on realistic
depth synthesis to bridge the sim-to-real gap for complex environments. 

\par

For navigation in diverse, large-scale environments, depth is also a critical input.
Platforms like Habitat~\cite{savva2019habitat} enable large-scale RL training for tasks such as
PointGoal Navigation, where agents frequently use depth or depth-derived features to achieve
near-perfect performance in previously unseen environments~\cite{Wijmans2019DDPPOLN}.
This necessity for robust visual priors also motivates auxiliary pretraining; for instance,
in~\cite{yang2025improving}, the authors introduce a spatially enhanced memory unit and demonstrate
improved sim-to-real transfer of the navigation policy by leveraging a frozen Variational Auto
Encoder (VAE) pretrained on a large-scale depth dataset~\cite{wang2020tartanair}.

\subsubsection{Manipulation and Grasping}

Depth perception is a critical modality for manipulation and contact-rich tasks in robotics.
Various approaches use depth images to construct explicit representations such as voxel
grids~\cite{james2022coarse}, point clouds~\cite{chen2023visualdex}, or object
meshes~\cite{pitz2024learning}, which are provided to the policy as input.
Although shown effective for manipulation, these methods often struggle with cluttered or dynamic
scenes and require careful calibration of perception pipelines.
To overcome these limitations, recent research has shifted towards end-to-end policies that directly
consume egocentric visual observations.
Zeng~\etal~\cite{zeng2018learning} propose a Q-Learning framework to train dense affordance maps on
visual RGB-D observations to clear cluttered scenes.
Kalashnikov~\etal~\cite{kalashnikov2018scalable} use continuous Q-learning on large-scale real
robot data to train general closed-loop grasping policies in a continuous action space.
However, both approaches control a two-fingered gripper with a simple binary open/close command.

Due to the high-dimensional control space of dexterous hands, recent
works~\cite{chen2023visualdex, lin2025sim, singh2024dextrah, he2025viral} adopt a teacher-student
training paradigm, where a teacher policy receives privileged proprioceptive information and is
distilled into a student policy that relies solely on visual observations. 
Singh~\etal~\cite{singh2024dextrah} train a student policy on synthetic stereo RGB images and employ
extensive domain randomization and image augmentation to handle visual sim-to-real gaps, which
results in long training times.
To further address the realization gap between teacher and student policies, their recent
work~\cite{singh2025end} provides the teacher with depth observations alongside privileged
proprioceptive inputs.
They demonstrate that distilling such a teacher into an RGB-based student results in more robust
manipulation policies.
In contrast, Zhang~\etal~\cite{zhang2025robustdexgrasp} train a teacher policy using the full point
cloud of the object and distill it to a student policy that receives only a partial point cloud of
the scene obtained from a single-view depth image.

\subsubsection{The Untapped Potential of Depth Foundation Models}

This extensive body of literature highlights two critical facts: 1) Depth is a popular modality for
effective sim-to-real and geometric control due to its invariance to light/texture and the ease of
high-fidelity simulation~\cite{mittal2025isaac, mittal2023orbit, tao2024maniskill3}; and 2)
geometric and semantic reasoning is crucial for success in core robotics tasks.
However, a critical gap remains: nearly all referenced RL and visual control methods relying on
depth use custom, task-specific encoders.
These encoders are typically trained from scratch alongside the policy, which is inefficient and
restricts the learned representation's generalization to the specific task and environmental domain
in which it was trained.
This highlights the lack of available depth encoders trained using a large-scale VFM pretraining
objective.
We argue that the same self-supervised scaling laws that revolutionized RGB vision
(Sec.~\ref{sec:lit_review_vfm}) can be applied to the depth modality.
By introducing DeFM, we aim to close this gap with, to the best of our knowledge, the first
large-scale depth foundation model for robotics, providing features that transfer across the
diverse locomotion, navigation, and manipulation tasks reviewed here.

%% file: chapters/3_prelimnary.tex
We build on DINOv2~\cite{oquab2023dinov2}, whose frozen features achieve strong performance on dense
prediction tasks such as depth estimation and semantic segmentation.
This makes it a suitable \ac{SSL} paradigm for transferring to the depth modality.

DINOv2 applies self-distillation to a \ac{ViT}: a student network is trained to match the output
distribution of a teacher network, whose weights are an exponential moving average of the student's.
At the image level, the DINO loss~\cite{caron2021emerging} is the cross-entropy between the
teacher's target distribution $p_t$ and the student's prediction $p_s$, both computed from the class
token of different augmented views of the same image:
\begin{equation}
    \mathcal{L}_{\text{DINO}} = - \sum p_t \log p_s .
\end{equation}
At the patch level, the iBOT loss~\cite{zhou2021ibot} applies the same cross-entropy to patch tokens,
where the student predicts the teacher's features for randomly masked patches, promoting dense
spatial correspondences.
Training is further stabilized by a KoLeo regularizer~\cite{sablayrolles2018spreading}, which
spreads features across the embedding space, and by Sinkhorn-Knopp centering of the teacher outputs.
For further details, we refer the reader to~\cite{oquab2023dinov2}.
The exact objective used for DeFM is described in Sec.~\ref{sec:overview}.

%% file: chapters/4_method.tex
\subsection{Overview}
\label{sec:overview}

\begin{figure*}[t]
    \centering
    \includegraphics[width=0.85\linewidth]{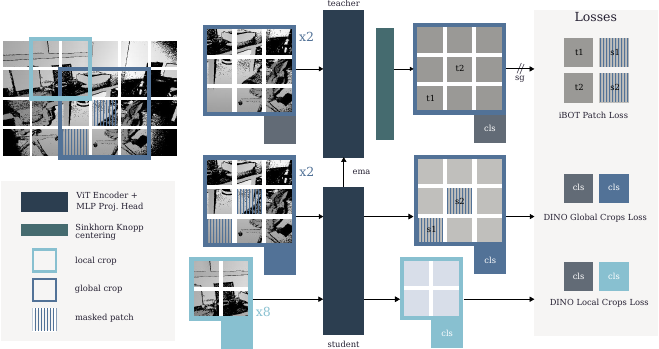}
    \caption{Overview of the self-supervised pretraining of DeFM.}
    \label{fig:overview_ssl}
\end{figure*}

We define a depth image as a 2D array in which each valid pixel represents the physical distance
along the camera's $Z$-axis, measured in meters.
To ensure consistency across varying sensor types and failure modes, any missing or invalid depth
pixels are strictly set to zero.
This standardized array acts as the base metric depth $D$ for our derived input representations.

Our core pretraining leverages the DINOv2 self-distillation framework (Fig. \ref{fig:overview_ssl}),
which we adapt for the depth modality through three key contributions.
First, to learn metric-aware features across diverse scales, we introduce a novel input
normalization strategy (Sec.~\ref{sec:normalization}).
Next, we train our largest model, a ViT-L with \SI{304}{M} parameters, on our curated depth dataset
(Sec.~\ref{sec:dataset}).
Lastly, we distill the knowledge from DeFM ViT-L into smaller, lightweight architectures (e.g.,
ViT-S, ResNets)  for resource-constrained robotic deployment (Sec.~\ref{sec:distillation}). 


Specifically, for a given input depth image $x$, we prepare $G$ large \textit{global crops}
($x_{g}$) and $L$ smaller \textit{local crops} ($x_{l}$) by applying various geometric and
photometric augmentations tailored for depth.
The \textit{global crops} are processed by the momentum-updated teacher network ($f_t$) and undergo
Sinkhorn-Knopp centering to produce the target distributions $p_{t}$.
The student network ($f_s$) processes the local crops and the partially masked global crops
($x'_{g}$).
Three primary losses are then applied to align the student's output with the teacher's targets:

\begin{itemize}
    \item DINO \textit{Global Crops} Loss ($\mathcal{L}_{\text{Global}}$): This loss aligns the
        student's representation of the partially masked global crops ($x'_{g}$) with the teacher's
        representation of the unmasked global crops ($x_{g}$).
        It operates on the $\mathtt{cls}$ token features:
    \begin{equation}
        \mathcal{L}_{\text{Global}} = \sum_{i=1}^{G} \sum_{j=1, j \neq i}^{G}
        \mathcal{L}_{\text{DINO}}(f_s(x'_{gi}), f_t(x_{gj})) 
    \end{equation}

    \item DINO \textit{Local Crops} Loss ($\mathcal{L}_{\text{Local}}$): This term aligns the
        student's representations of the local crops ($x_{l}$) with the teacher's representations of the
        global crops ($x_{g}$).
        This is computed between the $\mathtt{cls}$ tokens:
    \begin{equation}
        \mathcal{L}_{\text{Local}} = \sum_{g=1}^{G} \sum_{l=1}^{L} 
        \mathcal{L}_{\text{DINO}}(f_s(x_l), f_t(x_g))
    \end{equation}

    \item iBOT \textit{Patch Loss} ($\mathcal{L}_{\text{iBOT}}$): This loss is essential for
        learning dense spatial features.
        It applies the cross-entropy loss between the student's feature prediction ($p_{si}$) for
        randomly masked input patches and the teacher's target distribution ($p_{ti}$) derived from the
        corresponding visible patches in the teacher network:
    \begin{equation}
        \mathcal{L}_{\text{iBOT}} = - \sum_{i \in \text{masked}} p_{ti} \log p_{si}
    \end{equation}
\end{itemize}
The total loss is the sum of these three terms, along with the KoLeo regularizer ($\alpha = 0.1$),

\begin{equation}
    \mathcal{L}_{\text{total}} = \mathcal{L}_{\text{Global}} +  \mathcal{L}_{\text{Local}} +
    \mathcal{L}_{\text{iBOT}} + \alpha \mathcal{L}_{\text{KoLeo}}.
\end{equation}

\subsection{Dataset}
\label{sec:dataset}

As shown in \cite{oquab2023dinov2}, the development of a strong foundation model necessitates both
scale and diversity in pretraining data.
For a depth-specific foundation model tailored for robotics, this need is crucial: the encoder must
learn robust geometric and semantic priors spanning a wide range of scales and tasks, from
centimeter-scale dexterous manipulation to large-scale outdoor navigation and autonomous driving. 

To address this challenge, we curated a dataset totaling \SI{60.4}{M} depth images, combining 18
distinct datasets (Tab. \ref{tab:datasets}).
This collection is broadly classified into three types based on the depth image source, ensuring
maximal coverage across sensor fidelity and domain:

\begin{itemize}
        \item Monocular Depth Estimation (MDE): While the RGB domain benefits from curated, large-scale
        object-centric datasets (e.g., ImageNet-21k~\cite{deng2009imagenet}), no equivalent diverse
        dataset exists for depth.
        To ensure DeFM learns strong object-centric priors for various entities, such as cups, fruits,
        tools, etc., we leverage these RGB datasets by converting them into depth images using an
        off-the-shelf MDE network.
        We use the relative-depth estimation model of~\cite{yang2024depth} and do not calibrate these 
        predictions to metric depth. 
        Instead, each prediction is converted into a pseudo-metric depth map with a fixed
        depth range.
        The MDE samples thus contribute object-centric diversity and relative geometric structure,
        while absolute metric scale is learned from the real and synthetic data.

    \item Synthetic: These datasets provide clean, noise-free metric depth images and offer domain
        diversity costly to replicate at scale in the real world.
    
    \item Real: This category provides the noise characteristics inherent in real robotic sensor
        data.
        By incorporating diverse real-world sources collected with different depth sensors, the model
        learns to be robust and invariant to common sensor artifacts, missing data, and noise profiles.
        Learning these noise-invariant features is critical for effective sim-to-real transfer in
        robotic deployments.
    
\end{itemize}

Overall, the mixture of these sources provides a balance in diversity, scale, and noise fidelity,
thereby maximizing the learned representation's ability to generalize across diverse, unknown
robotic environments.

\begin{table}[t]
\centering
\caption{Overview of depth pretraining dataset collection. MDE denotes monocular depth estimation.}
\label{tab:datasets}
\renewcommand{\arraystretch}{1.1}
\setlength{\tabcolsep}{6pt}
\begin{tabular}{lrrr}
\toprule
\textbf{Dataset Name} & \textbf{Type} & \textbf{Domain} & \textbf{Size (M)} \\
\midrule
ImageNet-21k~\cite{deng2009imagenet} & MDE & Objects & 14.0 \\
SA-1B~\cite{kirillov2023segment} & MDE & Objects, Indoor, Outdoor & 11.0 \\
\midrule
Replica~\cite{straub2019replica} & Synthetic & Indoor & 0.3 \\
Hypersim~\cite{roberts2021hypersim} & Synthetic & Indoor & 0.07 \\
Meta GraspNet~\cite{gilles2023metagraspnetv2} & Synthetic & Manipulation & 0.3 \\
TartanGround~\cite{patel2025tartanground} & Synthetic & Indoor, Outdoor & 5.6 \\
TartanAir~\cite{wang2020tartanair} & Synthetic & Indoor, Outdoor & 5.8 \\
SHIFT~\cite{sun2022shift} & Synthetic & Autonomous Driving & 0.9 \\
\midrule
Taskonomy~\cite{zamir2018taskonomy} & Real & Indoor & 4.0 \\
HM3D~\cite{ramakrishnan2021habitat} & Real & Indoor & 11.0 \\
ARKitScenes~\cite{baruch2021arkitscenes} & Real & Indoor & 3.0 \\
ScanNet~\cite{dai2017scannet} & Real & Indoor & 0.4 \\
GraspNet-1B~\cite{fang2020graspnet} & Real & Manipulation & 0.1 \\
SUN3D~\cite{xiao2013sun3d} & Real & Indoor & 0.5 \\
DROID~\cite{khazatsky2024droid} & Real & Manipulation & 2.8 \\
GrandTour~\cite{frey2025boxi} & Real & Indoor, Outdoor & 0.2 \\
DIML~\cite{cho2021diml} & Real & Outdoor & 0.2 \\
DrivingStereo~\cite{yang2019drivingstereo} & Real & Autonomous Driving & 0.18 \\
\midrule
\textbf{Total} & & & \textbf{60.4} \\
\bottomrule
\end{tabular}
\vspace{-5mm}
\end{table}

\subsection{Input Normalization}
\label{sec:normalization}

\begin{figure*}
    \centering
    \includegraphics[width=\linewidth]{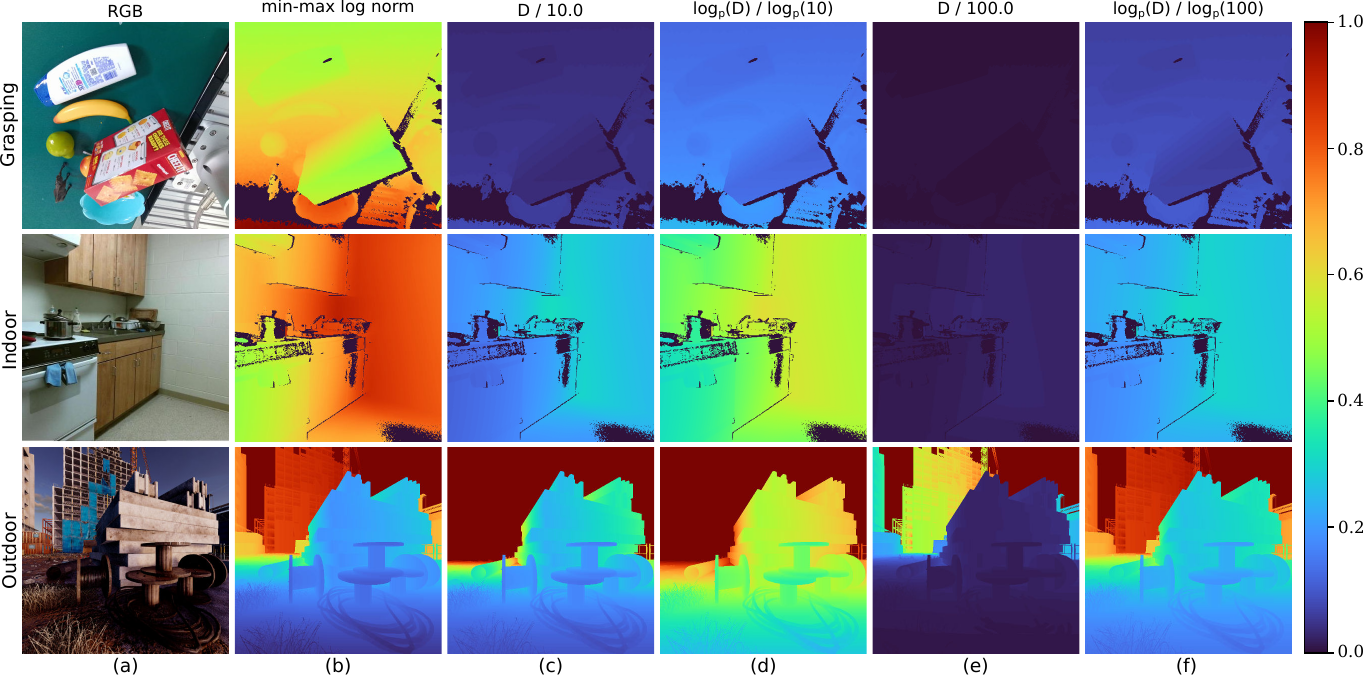}
    \caption{Visualization of different depth input normalization methods.
    Log normalization effectively captures the overall depth range while preserving fine-grained
    structure in near-field regions (b, d, f).
    In contrast, standard metric normalization yields weaker contrast and poorly separated gradients
    (c, e).
    In our representation, we stack columns (b), (d), and (f) to form a 3-channel normalized depth
    input, which retains metric awareness across diverse domains.}
    \label{fig:input_normalization}
\end{figure*}

A depth foundation model must operate effectively across an exceptionally wide range of depth
scales, from millimeter-level precision required in dexterous manipulation to tens or even hundreds
of meters in outdoor navigation.
Such a model must have metric awareness while attending to fine-grained details in the scene.
In practice, however, near-field depth variations matter disproportionately more for robotic
decision-making (e.g., avoiding an obstacle at \SI{1}{m}) than those in the far-field.
This observation motivates the use of logarithmic depth compression, which enhances sensitivity to
close-range structure while smoothly compressing large depth values, a strategy widely employed in
autonomous driving and depth-estimation systems~\cite{fu2018deep}.



Rather than choosing a single input normalization strategy, we thus propose a three-channel
log-compressed depth representation to capture depth structure across all operational regimes while
retaining metric awareness.
Let
\[
\log_p(D) = \log(1 + D)
\]
denote the standard \texttt{log1p} transform.
From the raw metric depth $D$, we construct three complementary channels:

\begin{enumerate}
    \item Global Log-Scaled Depth:
    This channel normalizes the log-compressed depth using the minimum ($D_{\min}$) and maximum
    ($D_{\max}$) depth within the current image, thereby preserving relative geometric structure
    (Fig.~\ref{fig:input_normalization}\nobreakdash-b):
    \[
    C_1 =
    \frac{\log_p(D) - \log_p(D_{\min})}
    {\log_p(D_{\max}) - \log_p(D_{\min})}.
    \]

    \item Mid-Range Norm:
    This channel emphasizes the depth regime most relevant for manipulation and indoor interaction
    (Fig. \ref{fig:input_normalization}-d):
    \[
    C_2 = \frac{\log_p(D)}{\log_p(10)}.
    \]

    \item Far-Range Norm:
    This channel emphasizes the depth regime most relevant for long-range navigation and outdoor
    scenes (Fig. \ref{fig:input_normalization}-f):
    \[
    C_3 = \frac{\log_p(D)}{\log_p(100)}.
    \]
\end{enumerate}

In summary, 2) and 3) retain metric awareness at different scales while 1) provides relative depth,
allowing for attending to finer details in close range, such as those required for manipulation. 
For the MDE-derived samples, which are not calibrated to metric depth, 
$C_2$ and $C_3$ still provide a consistent monotonic depth encoding but carry no physical scale,
whereas the scale-invariant $C_1$ remains equally valid.
The three channels are stacked to form the final depth input representation:
\[
X_{\text{in}} = [C_1, C_2, C_3].
\]

Finally, we apply a global mean and standard deviation normalization, computed once across the
entire pretraining dataset.
As visualized in Fig.~\ref{fig:input_normalization}, our representation preserves global metric
awareness, maintains fine-grained near-field structure, and provides stable gradients, yielding a
scalable depth representation for self-supervised pretraining.

\subsection{Implementation Details}

We train DeFM using the Fully-Sharded Data Parallel (FSDP) implementation of
DINOv2~\cite{oquab2023dinov2} and a ViT-L/14 backbone.
We prepare two global crops at $224^2$ resolution and eight local crops at $98^2$.
As in \cite{oquab2023dinov2}, we use separate MLP projection heads for DINO and iBOT losses.
Training was conducted for $625k$ iterations on 96 NVIDIA GH200 Grace Hopper GPUs using a
distributed batch size of 3,072.
We use AdamW optimizer with a peak learning rate of $3.5 \times 10^{-4}$ and a cosine weight decay
schedule (\SI{0.04}{} to \SI{0.2}{}).
We apply a $100k$ iteration learning rate warmup and update the teacher network's momentum using a
cosine schedule (0.994 to 1.0).
All training is performed in float16 mixed precision.
Due to computational constraints, we adopt the DINOv2 training recipe without tuning, including the
objective, projection heads, masking strategy, multi-crop setup, and optimization schedules.
The modifications are confined to the pretraining data (Sec.~\ref{sec:dataset}), the input
representation (Sec.~\ref{sec:normalization}), and the augmentations. The RGB-specific color
jittering, grayscale conversion, and solarization are replaced by depth-specific intensity jitter,
Gaussian noise, and random pixel dropout simulating missing sensor depth.
The full pretraining hyperparameters are listed in
Appendix~\ref{app:impl}.

\subsection{Distillation}
\label{sec:distillation}

\begin{figure*}
    \centering
    \includegraphics[width=0.9\linewidth]{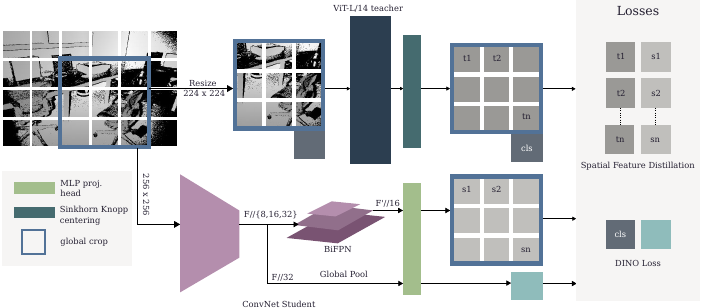}
    \caption{Overview of distilling the DeFM-L/14 teacher into CNN backbones.
    A BiFPN module is added on top of the CNN encoder to produce dense spatial features, which are
    supervised using the teacher’s spatial tokens.
    The teacher’s class token provides global supervision to the CNN’s pooled feature
    representation.}
    \label{fig:cnn_distillation}
\end{figure*}

Knowledge distillation~\cite{hinton2015distilling} aims to transfer the knowledge of a large teacher
model into a smaller, more efficient student model by minimizing the distance between their outputs.
Since our self-supervised objective is fundamentally a distillation process from the teacher network
to the student, we leverage the same training loop with specific modifications to the student
network architectures.
We use DeFM-L/14 as a frozen teacher for all distillation runs and maintain a separate EMA of the
student’s weights, which serves as the final distilled model.

\subsubsection{Motivation and Target Architectures}

For many downstream RL tasks, the ability to use large batch sizes is critical for stable policy
optimization, such as in PPO~\cite{schulman2017proximal}.
Therefore, there is a need for smaller, efficient encoders to minimize memory
footprint~\cite{agarwal2023legged, singh2025end, yang2025improving} and faster training times.
We thus distill DeFM's representations into a ViT-S and three lightweight CNN families, namely,
ResNets~\cite{he2016deep}, RegNets~\cite{radosavovic2020designing}, and
EfficientNets~\cite{tan2019efficientnet}, ranging in size from \SI{3}{M} to \SI{30}{M} parameters.

\subsubsection{Architectural Adaptation via BiFPN}

Distilling global and dense spatial features from a ViT teacher into a standard CNN student is
non-trivial, as typical CNN last-layer global pooling discards the spatial information needed for
dense prediction tasks.
To ensure the CNN students retain the necessary dense spatial features, we integrate a \ac{BiFPN}
\cite{tan2020efficientdet} on top of our CNN encoders.
The \ac{BiFPN} performs bi-directional fusion of feature maps extracted from the CNN backbone at
three distinct resolutions ($F/\{8,16,32\}$).
The BiFPN then outputs enhanced, dense spatial features at the same resolutions. 

\subsubsection{Distillation Strategy and Hyperparameters}

We utilize the same crop strategy as the original pretraining.
However, we adjust the input size to align the student and teacher feature maps: we pass global
crops of size $256 \times 256$ to the CNN students.
This ensures the output spatial feature map from the BiFPN at the $F'/16$ resolution (i.e.,
$16 \times 16$) aligns with the spatial tokens of the ViT-L/14 teacher
($224/14 \approx 16 \times 16$).

\begin{itemize}
    \item {Global Feature Distillation ($\mathcal{L}_{\text{DINO}}$):} We match the globally pooled
        feature map of the CNN encoder with the $\mathtt{cls}$ token of the frozen ViT teacher.
    \item {Spatial Feature Distillation ($\mathcal{L}_{\text{iBOT}}$):} We remove the masking and
        match the dense spatial output of the \ac{BiFPN} at the $F'/16$ resolution with the spatial
        patch tokens of the ViT teacher for both global crops.
\end{itemize}


As before, separate MLP projection heads are used for both the DINO and spatial feature losses.
We use a multi-student distillation approach for computational efficiency~\cite{simeoni2025dinov3}.
Distillation was run for $200k$ iterations on $64$ NVIDIA GH200s with a batch size of $2,048$ and a
learning rate of $1 \times 10^{-3}$ for each student.

%% file: chapters/5_experiments.tex
In this section, we aim to evaluate the quality and generalizability of the representations learned
by DeFM.
For a model to be considered truly foundational, its learned features must successfully transfer
across different tasks, domains, and sensors.
We first provide a qualitative assessment by performing \ac{PCA} on the extracted features
(Sec.~\ref{sec:pca}) to visualize the emergent feature structure.
We then quantitatively assess our frozen encoder's transfer capability across diverse benchmarks.
This includes evaluation via linear probing for classification (Sec.~\ref{sec:classification}) and
evaluation on the dense prediction task of semantic segmentation (Sec.~\ref{sec:segmentation}).
Next, we present ablations for our design choices, including the input normalization strategy, the
pretraining data composition, and the BiFPN architecture for CNN distillations
(Sec.~\ref{sec:ablations}).
Lastly, as our focus is on robotic applications, we compare the inference times of all our DeFM and
distilled models in Sec.~\ref{sec:inference}.
We discuss in detail the performance of DeFM for various robotics tasks in a separate dedicated
Section (Sec.~\ref{sec:Robotic_experiments}).

\subsection{Foundational Depth Image Understanding}
\label{sec:pca}

\begin{figure}[t]
    \centering
    \includegraphics[width=\linewidth]{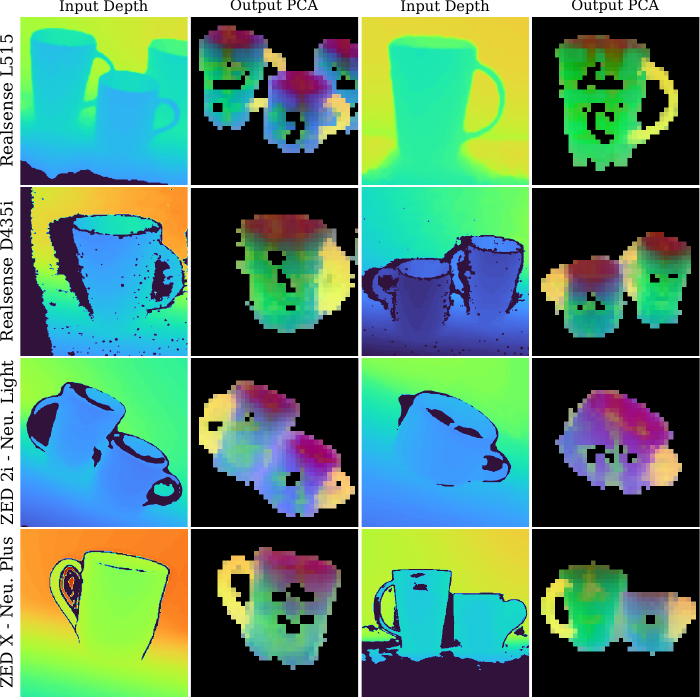}
    \caption{PCA visualization of the patch features obtained from the DeFM-L/14 encoder when
    processing depth images of various cups captured by different sensors.
    The first three PCA components are mapped to the RGB color channels for visualization.
    Notice the feature consistency of the cup handle (visualized in yellow) across all images,
    demonstrating that DeFM learns a useful prior for a robotic grasping task.
    The background is removed by thresholding the first PCA component.}
    \label{fig:pca_cups}
    \vspace{-5mm}
\end{figure}

While depth images provide very strong geometric priors, we argue that meaningful semantic features
can also be obtained from them despite the absence of texture and color.
To demonstrate this, we perform PCA on features extracted by our DeFM-L/14 encoder.
\par

We collect depth images of various cups using four different depth sensors, namely, RealSense L515
(solid state LiDAR depth camera), RealSense D435i (stereo infrared), ZED 2i (with neural Light
depth) and ZED X (with neural Plus depth).
We then fit a PCA on the extracted patch features and visualize the first three components, mapping
them to RGB channels (Fig. \ref{fig:pca_cups}).
We observe a strong, consistent correlation between the features and the functional parts of the cup
across all sensors.
For example, the cup handle is consistently clustered and visualized in yellow, while the cup rim is
associated with magenta.
The consistency of these semantic features across diverse sensor modalities has significant
implications for robotics, serving as a robust prior on where to grasp for downstream policy
learning.
\par

In Fig.~\ref{fig:highlight}-V, the model successfully differentiates and highlights the handles
across various types of drawers and cabinets.
Overall, these qualitative experiments suggest that DeFM learns consistent, part-level features
across various object classes and depth sensor types.

\subsection{Classification}
\label{sec:classification}

The classification task evaluates the quality of the learned visual representations by testing their
ability to distinguish between different object categories.
Due to the unavailability of a standard classification benchmark for depth images, we create an
ImageNet-1k-Depth benchmark consisting of RGB converted depth images estimated from an MDE
network~\cite{yang2024depth}. As in DINOv2, this benchmark measures the quality of the learned
representation on the pretraining domain.

Linear probing is the standard evaluation protocol for \acp{VFM}.
In this setup, the pretrained encoder is kept frozen, and only a lightweight linear classifier is
trained on labeled data.
Strong linear-probe performance indicates that the encoder has learned a feature space where classes
are linearly separable, reflecting strong generalization without task-specific fine-tuning.
We additionally include top-1 and top-5 KNN accuracy, which evaluates how well the frozen embeddings
cluster by retrieving labels from their nearest neighbors in feature space.
This training-free evaluation provides a complementary and typically lower-bound estimate of
representation quality relative to linear probing.
The full classification probing protocol is given in Appendix~\ref{app:impl}
(Tab.~\ref{tab:app_probe}).

\begin{table}[t]
\centering
\caption{Classification evaluations on ImageNet-1k-Depth.}
\label{tab:knn_linear_results}
\renewcommand{\arraystretch}{1.2}
\setlength{\tabcolsep}{5pt}
\resizebox{\columnwidth}{!}{%
\begin{tabular}{@{}llccc@{}}
\toprule
{Architecture} & {Feature} & {Top-1 KNN (↑)} & {Top-5 KNN (↑)} & {Linear (↑)} \\ 
\midrule
ViT-L/14  & DINOv2      & 60.28 & 80.75 & 68.26 \\
ViT-L/16 (CPE) & C-RADIOv3 & 55.23 & 76.96 & 63.92 \\
ViT-L/16  & DINOv3      & \textbf{64.07} & 83.48 & 69.55 \\
ViT-L/14  & DeFM & 63.46 & \textbf{84.79} & \textbf{71.72} \\ 
\midrule
DeiT-S/16 & Theia       & 28.17 & 47.47 & 33.14 \\
ViT-S/14  & DINOv2      & 47.66 & 69.11 & 55.06 \\
ViT-S/16  & DINOv3      & 49.50 & 70.93 & 50.86 \\
ViT-S/14  & DeFM & \textbf{55.27} & \textbf{78.06} & \textbf{61.54} \\
\bottomrule
\end{tabular}%
}
\end{table}

\begin{table}[t]
\centering
\caption{Comparison of distilled CNN architectures on ImageNet-1k-Depth Classification.}
\label{tab:architecture_results}
\renewcommand{\arraystretch}{1.1}
\setlength{\tabcolsep}{6pt}
\resizebox{\columnwidth}{!}{%
\begin{tabular}{@{}lcccc@{}}
\toprule
{Model} & {Params (M)} & {Top-1 KNN (↑)} & {Top-5 KNN (↑)} & {Linear (↑)} \\
\midrule
ViT-S/14 & 21.6 & 55.27 & 78.06 & 61.54 \\
ViT-L/14 & 303.7 & 63.46 & 84.79 & 71.72 \\
\addlinespace[0.2em]
\midrule
\addlinespace[0.2em]
ResNet-18 & 11.7 & 45.71 & 69.69 & 50.58 \\
ResNet-34 & 23.3 & 48.16 & 72.72 & 54.39 \\
ResNet-50 & 26.2 & 53.71 & 77.63 & 61.54 \\
\addlinespace[0.2em]
\midrule
\addlinespace[0.2em]
RegNetY-400MF & 4.1 & 49.51 & 72.87 & 50.51 \\
RegNetY-800MF & 6.3 & 51.58 & 74.91 & 57.03 \\
RegNetY-1.6GF & 12.4 & 53.03 & 76.21 & 57.28 \\
\addlinespace[0.2em]
\midrule
\addlinespace[0.2em]
EfficientNet-B0 & 3.01 & 44.38 & 67.98 & 46.17 \\
EfficientNet-B2 & 4.95 & 47.43 & 71.51 & 50.32 \\
EfficientNet-B4 & 14.16 & 50.33 & 74.74 & 54.73 \\
EfficientNet-B6 & 28.98 & 53.35 & 77.81 & 59.23 \\
\bottomrule
\end{tabular}%
}
\end{table}

\subsubsection{Baselines}
\label{sec:baselines}
We compare the performance of our models against state-of-the-art \acp{VFM} from the RGB domain
(inferring on depth images) that represent different pretraining paradigms:

\begin{itemize}
    \item ViT-L comparison: We compare our largest model, DeFM-L/14, against the self-distilled
        VFMs: DINOv2~\cite{oquab2023dinov2} and the more recent DINOv3~\cite{simeoni2025dinov3}.
        We also include C-RADIOv3~\cite{heinrich2025radiov25improvedbaselinesagglomerative}, an
        agglomerative model that distills features from multiple VFMs into a single encoder.
    \item ViT-S comparison: For the smaller, resource-efficient model, DeFM-S/14, we compare against
        the corresponding size variants of DINOv2 and DINOv3.
        Since C-RADIOv3 is not available in this small size, we instead use Theia-S
        \cite{shang2024theia}, which is another agglomerative model specifically distilled for robotics
        tasks.
\end{itemize}

For all RGB-pretrained baselines, we perform min-max normalization on the depth image, stack it into
three channels, and finally normalize using the standard ImageNet mean and standard deviation.
We found this standard approach performs better than our depth-specific normalization strategy for
these RGB-pretrained models.

\subsubsection{Results}

The quantitative results (Tab.~\ref{tab:knn_linear_results}) demonstrate the superior feature
quality of our DeFM encoder:

\begin{itemize}
    \item ViT-L performance: DeFM-L/14 achieves the highest score on both the Top-5 KNN (84.79\%)
        and linear probing (71.72\%) metrics, while remaining highly competitive with the DINOv3 model
        on the Top-1 KNN metric.
    \item ViT-S performance: DeFM-S/14 outperforms existing state-of-the-art models in its size
        category by up to $10\%$ across all three metrics.
        This large performance gap highlights the limitations of existing smaller distilled RGB models
        when applied to depth and strongly reinforces the need for smaller-sized DeFM variants for
        efficient, high-performance robotic deployments.
    
\end{itemize}

Surprisingly, employing depth information alone yields a 71.7\% accuracy on a 1000-class object
classification task.
While this performance is lower than the current state-of-the-art models for RGB input, which
achieve around $87.2\%$ accuracy~\cite{simeoni2025dinov3}, it clearly demonstrates the intrinsic
power of depth data.
The result is particularly impressive, given that the classification task includes challenges such
as differentiating between various snake species or fine-grained variations in furniture, tasks that
typically require human experts to rely on color and texture cues.




\subsubsection{Performance of CNNs}

We evaluate all of our distilled CNN variants of DeFM to understand the tradeoff between compute
time and feature quality, as these smaller models are essential for efficient robotics deployment.
The results are summarized in Tab.~\ref{tab:architecture_results}.
As expected, we observe a general drop in performance correlating with decreasing model size. 
However, we note a surprisingly strong result: some of the very small CNN models, such as
RegNetY-800MF (6.3 M), outperform the ViT-S RGB baselines (22.1 M) (Tab.
\ref{tab:knn_linear_results}) across all three metrics.


\subsection{Semantic Segmentation}
\label{sec:segmentation}

Semantic segmentation is a critical evaluation task for VFMs, as it directly assesses the quality of
dense spatial features, which are essential for most robotic tasks, such as grasp localization,
obstacle avoidance, and scene understanding.
To assess whether the frozen DeFM encoder provides meaningful per-pixel representations, we perform
linear-probe segmentation.
Following the standard protocol, we train a linear layer to predict the class logits from the patch
tokens.
The output is then upsampled to the full resolution using bilinear interpolation to obtain the final
segmentation map.
The full segmentation probing protocol and dataset-specific settings are given in
Appendix~\ref{app:impl} (Tab.~\ref{tab:app_seg}).
We employ the same baselines introduced earlier for classification in Sec.~\ref{sec:baselines}.
In addition, we compare against Depth Anything v2 (DAv2)~\cite{yang2024depth}, an MDE FM known for its
strong geometric representations, as well as two recent point cloud FMs,
Concerto~\cite{zhang2026concerto} and Utonia~\cite{zhang2026utonia}.
To evaluate the point cloud baselines, we unproject the 2D depth images into 3D space using the
camera intrinsics, extract features through the frozen point cloud encoders, and perform standard
linear probing.
\subsubsection{Datasets}

An ideal foundation model must demonstrate robust generalization across diverse domains and sensor
types.
Accordingly, we select 5 diverse depth datasets for evaluating semantic segmentation, ensuring
coverage across varied environments and acquisition modalities.
These benchmarks range from indoor room settings and fine-grained tabletop manipulation scenarios to
large-scale outdoor navigation environments.
The datasets have been summarized in Tab. \ref{tab:seg_datasets}. 

\begin{table*}[t]
\centering
\caption{Summary of segmentation datasets used for downstream evaluation.}
\label{tab:seg_datasets}
\renewcommand{\arraystretch}{1.1}
\setlength{\tabcolsep}{4.5pt}
\begin{tabular}{@{}llccc@{}}
\toprule
{Dataset} & {Domain / Scene Type} & {\# Classes} & {Range (m)} & {Sensor} \\ 
\midrule
ScanNet~\cite{dai2017scannet} & Indoor room scenes & 40 & $<10$ & Structure Sensor \\
SUN-RGBD~\cite{song2015sun} & Indoor rooms / building scenes & 37 & $<10$ & Kinect v1-v2 / RealSense
/ Xtion \\
TartanGround~\cite{patel2025tartanground} & Outdoor construction, industry, infrastructure & 6 &
$<100$ & Simulated \\
OFFSED~\cite{neigel2021offsed} & Outdoor off-road environments & 5 & $<30$ & ZED Stereo Camera \\
GraspNet-1B~\cite{fang2020graspnet} & Tabletop object grasping & 40 & $<1$ & Azure Kinect \\ 
\bottomrule
\end{tabular}%
\end{table*}

\subsubsection{Results}

\begin{figure*}[t]
    \centering
    \includegraphics[width=\linewidth]{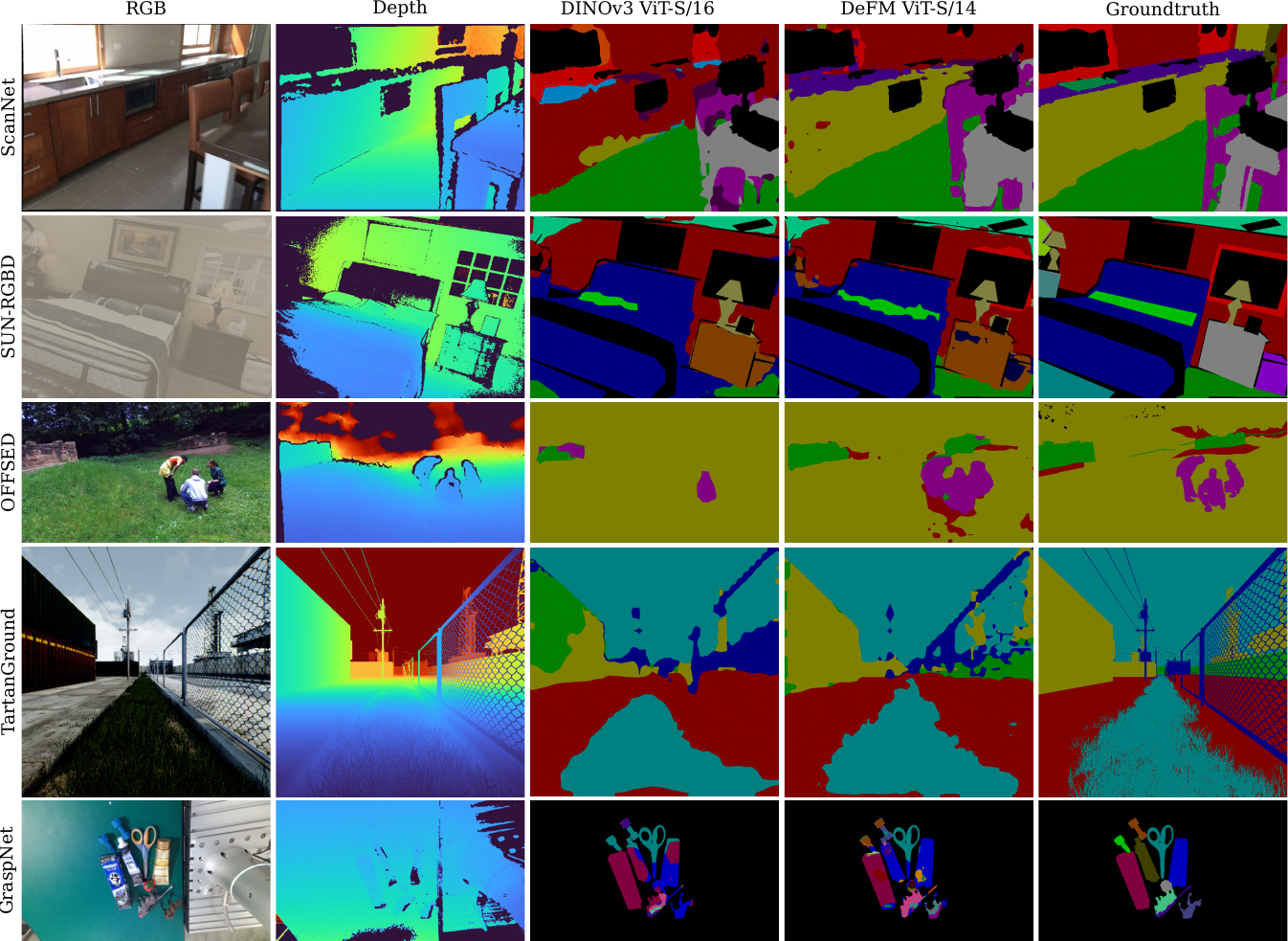}
    \caption{Qualitative results of semantic segmentation on different datasets.}
    \label{fig:segmentation_viz}
\end{figure*}

The quantitative results in terms of mean Intersection over Union (mIoU) are summarized in
Tab.~\ref{tab:seg_results}.
Overall, DeFM models demonstrate robust generalization across different domains and surpass existing
baselines in the majority of benchmarks.
For the ViT-L architecture, DeFM-L/14 achieves the highest mIoU scores across four out of five
datasets, while the performance improvement is particularly pronounced for the ViT-S architecture.
Consistent with our classification findings, the generalization gap is significant, with DeFM-S/14
achieving mIoU scores up to $30\%$ higher than the corresponding baselines in several benchmarks.
We illustrate the improvement qualitatively by comparing the segmentation output of DINOv3-S/16 with
our DeFM-S/14 in Fig.~\ref{fig:segmentation_viz}.
\par

Furthermore, our smaller distilled ResNet models consistently outperform the similar-sized baseline
models on most datasets, providing strong geometric and semantic features for
resource-constrained robotic deployments.
Finally, point cloud FMs perform worse than even our smallest DeFM models.
This is expected because partial, camera-based point clouds represent a major domain shift from the
pretraining datasets of Utonia and Concerto, which primarily rely on dense room scans, isolated
objects, and LiDAR scenes.

\begin{table*}[t]
\centering
\caption{Semantic Segmentation evaluation on various datasets (mIoU) (↑).}
\label{tab:seg_results}
\renewcommand{\arraystretch}{1.2}
\setlength{\tabcolsep}{5pt}
\begin{tabular}{@{}llcccccc@{}}
\toprule
\multirow{2}{*}{{Model}} &
\multirow{2}{*}{{Pretraining}} &
\multicolumn{2}{c}{{Indoor}} &
\multicolumn{2}{c}{{Outdoor}} &
\multicolumn{1}{c}{{Manipulation}} \\
\cmidrule(lr){3-4} \cmidrule(lr){5-6} \cmidrule(lr){7-7}
& & {ScanNet} & {SUN-RGBD} & {OFFSED} & {TartanGround} & {GraspNet-1B} \\
\midrule
ViT-L/14 & DINOv2 & 29.91 & 32.71 & 53.98 & 65.88 & 24.12 \\
ViT-L/16 (CPE) & C-RADIOv3 & 31.17 & 33.83 & 55.40 & 70.46 & 24.65 \\
ViT-L/16 & DINOv3 & 34.86 & \textbf{39.23} & 54.32 & 68.61 & 25.52 \\
ViT-L/14 & DAv2 & 29.91 & 32.68 & 55.87 & 67.25 & 25.43 \\
ViT-L/14 & DeFM & \textbf{38.26} & 36.88 & \textbf{57.15} & \textbf{73.22} & \textbf{27.57} \\
\midrule
DeiT-S/16 & Theia & 19.56 & 17.48 & 41.56 & 55.13 & 11.11 \\
ViT-S/14 & DINOv2 & 22.90 & 23.77 & 48.06 & 61.83 & 15.92 \\
ViT-S/16 & DINOv3 & 26.31 & 26.55 & 56.08 & 63.59 & 18.04 \\
ViT-S/14 & DAv2 & 22.28 & 22.87 & 47.63 & 63.98 & 14.22 \\
ViT-S/14 & DeFM & \textbf{33.79} & \textbf{33.27} & \textbf{58.10} & \textbf{70.82} & \textbf{20.53}
\\
\midrule
ResNet-50 & DeFM & 35.34 & 32.19 & 52.55 & 70.89 & 31.95 \\
ResNet-34 & DeFM & 33.87 & 30.36 & 52.21 & 70.01 & 27.86 \\
ResNet-18 & DeFM & 28.01 & 25.97 & 46.87 & 65.49 & 23.27 \\
\midrule
PTv3-L (208M) & Concerto & 21.52 & 16.18 & 32.25 & 58.35 & 4.44 \\
RE-PTv3 (137M) & Utonia & 23.06 & 20.70 & 30.36 & 59.79 & 4.34 \\
\bottomrule
\end{tabular}%
\end{table*}

\subsection{Ablations}
\label{sec:ablations}

\begin{table}[t]
\centering
\caption{Ablation studies on input normalization, pretraining data, and CNN distillation
architectures.
Performance is reported as Semantic Segmentation mIoU (↑).}
\label{tab:ablations}
\renewcommand{\arraystretch}{1.2}
\setlength{\tabcolsep}{2pt}
\resizebox{\columnwidth}{!}{%
\begin{tabular}{@{}lcccccc@{}}
\toprule
\multirow{2}{*}{{Ablation}} &
\multicolumn{2}{c}{{Indoor}} &
\multicolumn{2}{c}{{Outdoor}} &
\multicolumn{1}{c}{{Manipulation}} \\
\cmidrule(lr){2-3} \cmidrule(lr){4-5} \cmidrule(lr){6-6}
& {ScanNet} & {SUN-RGBD} & {OFFSED} & {TartanGround} & {GraspNet-1B} \\
\midrule
\multicolumn{6}{@{}l}{\textit{Input Normalization }} \\
\midrule
$C_1$ (Global Log-Scaled) & 20.83 & 17.81 & 50.01 & 65.39 & 16.68 \\
$C_3$ (Far-Range Norm) & 24.23 & \textbf{20.93} & 51.63 & 65.76 & 15.38 \\
$[C_1, C_2, C_3]$ (Ours) & \textbf{24.43} & 20.76 & \textbf{52.06} & \textbf{67.11} & \textbf{17.07}
\\
\midrule
\multicolumn{6}{@{}l}{\textit{Pretraining Data}} \\
\midrule
Real only & \textbf{27.80} & \textbf{22.62} & 49.20 & 65.01 & 15.06 \\
Synthetic only & 21.75 & 18.66 & 50.23 & \textbf{67.14} & 15.31 \\
MDE only & 22.74 & 18.93 & 49.78 & 65.87 & 14.12 \\
All (Ours) & 24.43 & 20.76 & \textbf{52.06} & 67.11 & \textbf{17.07} \\
\midrule
\multicolumn{6}{@{}l}{\textit{CNN Distillation Architecture}} \\
\midrule
No FPN & 14.81 & 12.96 & 40.50 & 57.75 & 18.30 \\
FPN & \textbf{17.18} & 15.46 & \textbf{45.81} & 58.93 & 16.88 \\
BiFPN (Ours) & 17.13 & \textbf{16.35} & 42.39 & \textbf{60.24} & \textbf{19.32} \\
\bottomrule
\end{tabular}
}
\vspace{-5mm}
\end{table}

\subsubsection{Input Normalization}

We train a ViT-S model for 125K steps (batch size 1024) on our DeFM dataset using three different
input normalization strategies: (i) Global Log-Scaled Depth ($C_1$), (ii) Far-Range Norm ($C_3$),
and (iii) our proposed 3-channel stacked representation ($[C_1, C_2, C_3]$).
Evaluating these pretrained models via semantic segmentation linear probing
(Tab.~\ref{tab:ablations}) reveals that our 3-channel strategy achieves the highest performance
on four of the five benchmarks, while $C_3$ is marginally better on SUN-RGBD.
This success can be attributed to the model's capacity to concurrently exploit relative geometric
structures and absolute metric priors at multiple scales.
\par
Crucially, while the metric-aware $C_3$ channel generally outperforms the relative $C_1$ channel, it
falls short on the GraspNet-1B benchmark.
This suggests that far-range normalization overly compresses the fine-grained, close-range
details required for dexterous manipulation.
By unifying these complementary channels, the multi-channel representation provides an overall
benefit across the various spatial ranges encountered in robotics.

\subsubsection{Pretraining Data}

To assess the contribution of each data source, we pretrain the same ViT-S model with the same
budget (125K steps, batch size 1024) on only the real, only the synthetic, or only the MDE-derived
portion of our dataset, and compare it against the full mixture (Tab.~\ref{tab:ablations}).
We observe that each single-source variant is competitive on some benchmarks but
drops on others (e.g., real-only on OFFSED and TartanGround, synthetic-only on ScanNet and
SUN-RGBD, and MDE-only on GraspNet-1B).
The full mixture achieves the highest average mIoU and is the only variant that is never the
weakest on any benchmark, suggesting that the three sources are complementary.
This reduced-scale experiment does not establish the optimal
mixture at full scale, but we expect the benefit of data diversity to become more pronounced at larger
scale, which remains to be verified.

\subsubsection{BiFPN Architecture}
To evaluate the necessity of spatial feature aggregation during CNN distillation, we distill our
DeFM-S/14 into three ResNet-18 variants: (i) without a feature pyramid (No FPN), (ii) with a
standard FPN, and (iii) with a BiFPN.
As shown in Tab.~\ref{tab:ablations} on the semantic segmentation linear probing task, incorporating
a multi-scale fusion module significantly improves performance.
While the BiFPN achieves the strongest overall results, the standard FPN remains competitive,
demonstrating that either multi-scale fusion mechanism represents a viable architectural design
choice.

\subsection{Inference Times}
\label{sec:inference}

To facilitate informed hardware planning, we report comprehensive metrics for all models in the DeFM
family in Tab. \ref{tab:inference_times}.
This evaluation provides a crucial guideline for understanding the tradeoff between training times,
memory consumption, and performance.

We report the inference latency and VRAM consumption on an NVIDIA RTX 4090 (using a large batch
size, $\text{BS}=128$), which is a typical GPU used to train downstream RL policies.
Furthermore, we report the inference time for a single image ($\text{BS}=1$) on an NVIDIA Jetson AGX
Orin, a representative platform for real-world robotic deployments.
We use images of size $ 224\times224$ for all reported numbers. 

Our results highlight that inference speed on the Orin edge hardware is highly dependent on the
model's architecture itself, rather than solely on parameter count: RegNetY models, which often
incorporate features like Squeeze-and-Excitation (SE) layers, and EfficientNet models, which rely on
depth-wise separable convolutions, show unexpectedly high latency on the Orin platform as these
operations are not always fully optimized in standard PyTorch execution.

It is important to note that all reported timings are measured using the standard PyTorch framework,
and no model runtime optimizations (e.g., TensorRT/ONNX conversion) were performed.
Utilizing such optimizations is expected to reduce the latency across all models, particularly on
the resource-constrained Orin platform.



\begin{table}[t]
\centering
\caption{Comparison of inference times of DeFM models on training and deployment hardware.}
\label{tab:inference_times}
\renewcommand{\arraystretch}{1.2}
\setlength{\tabcolsep}{4pt}
\resizebox{\columnwidth}{!}{%
\begin{tabular}{@{}lrrrrr@{}}
\toprule
{Model} & \begin{tabular}[c]{@{}c@{}}{Params} \\ {(M)}\end{tabular} &
\begin{tabular}[c]{@{}c@{}}{FLOPs} \\ {(GFLOPs)}\end{tabular} & \begin{tabular}[c]{@{}c@{}}{RTX
4090} \\ {(ms)}\end{tabular} & \begin{tabular}[c]{@{}c@{}}{Memory} \\ {(GB)}\end{tabular} &
\begin{tabular}[c]{@{}c@{}}{Jetson AGX} \\ {Orin (ms)}\end{tabular} \\
\midrule
Batch Size & - & 128 & 128 & 128 & 1 \\
\midrule
ViT-S/14 & 21.6 & 707.48 & 63.76 & 0.99 & 11.92 \\
ViT-L/14 & 303.7 & 9962.24 & 624.91 & 3.32 & 72.82 \\
\midrule
ResNet-18 & 11.7 & 256.13 & 21.06 & 0.93 & 8.67 \\
ResNet-34 & 23.3 & 493.69 & 33.08 & 1.33 & 13.54 \\
ResNet-50 & 26.2 & 631.19 & 69.39 & 1.66 & 17.79 \\
\midrule
RegNetY-400MF & 4.1 & 60.05 & 17.27 & 1.07 & 25.17 \\
RegNetY-800MF & 6.3 & 133.28 & 25.21 & 1.56 & 24.16 \\
RegNetY-1.6GF & 12.4 & 296.20 & 44.25 & 1.39 & 41.82 \\
\midrule
EfficientNet-B0 & 3.01 & 52.72 & 29.39 & 1.85 & 21.04 \\
EfficientNet-B2 & 4.95 & 94.13 & 46.12 & 2.26 & 28.37 \\
EfficientNet-B4 & 14.16 & 256.24 & 86.51 & 3.50 & 39.67 \\
EfficientNet-B6 & 28.98 & 460.02 & 150.98 & 4.56 & 54.11 \\
\bottomrule
\end{tabular}%
}
\end{table}









%% file: chapters/6_robotic_experiments.tex
While our analysis in Sec.~\ref{sec:Experiments} established the strong generalization of DeFM's
frozen features on standard perception tasks (classification and segmentation), the main motivation
of this work is to provide representations that are directly usable for RL, ideally without any
task-specific fine-tuning.
Solving complex, embodied control tasks requires features that are not just semantically meaningful
but are geometrically grounded and action-relevant.
\par

To evaluate the quality of our DeFM representations on downstream tasks, we select three distinct
and challenging RL tasks across navigation, manipulation, and locomotion.
Specifically, our evaluation includes: 

\begin{itemize}
    \item \textbf{Embodiment Aware Point-Goal Nav} (Sec.~\ref{sec:sru_nav}): Utilizing the Unitree
        B2W wheeled-legged robot, this long-range task~\cite{yang2025improving} tests navigation
        policies that must handle the robot's kinematics and physical footprint, prioritizing accurate
        geometric obstacle avoidance.
    \item \textbf{Dexterous Grasping} (Sec.~\ref{sec:dex_grasping}): Using a KUKA-Allegro arm-hand
        setup, the agent is tasked with grasping a diverse range of objects~\cite{singh2024dextrah}.
    \item \textbf{Quadrupedal Ladder Climbing} (Sec.~\ref{sec:ladder_climbing}): This challenging
        locomotion task requires the ANYbotics ANYmal quadruped to robustly climb ladders of varying
        sizes and angles~\cite{vogel2024robust}.
        The policy relies on proprioception and features derived from four depth cameras to ensure
        stable footholds and gait synchronization.
\end{itemize}

\textbf{Baselines:} For each of the tasks, we compare our DeFM encoder with the following baselines:
\begin{enumerate}
\item ImageNet: We use the default ImageNet RGB pretrained weights for the encoder while using the
    3-channel stacking with min-max normalization as done earlier for the RGB pretrained baselines.
\item Variational Auto Encoder (VAE): Following the method in \cite{yang2025improving}, we pretrain
    a VAE using reconstruction loss on the DeFM dataset, along with using our 3-channel input
    normalization method.
\item DINOv3: We distill DINOv3-L features into the encoder on the DeFM dataset, while using the
    normalization for RGB pretrained baselines.
\end{enumerate}

In the subsequent subsections, we discuss in detail the setup and evaluations for each of the tasks.
Unless otherwise stated, we always use the frozen encoder and only train the downstream policy
network.

\subsection{Navigation: Embodiment Aware Point-Goal Nav}
\label{sec:sru_nav}

To test the suitability of DeFM features for navigation, we select the point goal navigation task
introduced in \cite{yang2025improving} and utilize the Unitree B2W wheeled legged robot.
In this task, the policy must perform long-range navigation over diverse terrains and complex
environments while actively avoiding obstacles using a single front-facing depth camera as input.

\subsubsection{Setup}

For encoding the depth images, a compact RegNetX-400MF encoder, along with a Feature Pyramid Network
(FPN), is used, which is kept frozen during the RL policy training~\cite{yang2025improving}.
The dense output of the FPN is then used to perform cross attention with the proprioceptive data and
processed by a Spatially-enhanced Recurrent Unit (SRU) block and a few MLP layers to generate the
final velocity commands, which are then tracked by a low-level locomotion policy.
For training the downstream RL policy, we adopt the hyperparameters specified in
\cite{yang2025improving} and train on a single \text{RTX 4090} GPU.
Downsampled depth images of size $64\times40$ are used as an input to the network.

\begin{figure}[t]
    \centering
    \includegraphics[width=\linewidth]{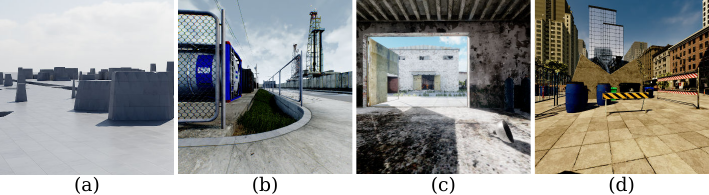}
    \caption{(a) Shows the training environment, while (b-d) represent the test environments for
    embodiment-aware navigation.}
    \label{fig:sru_sim_env}
    \vspace{-5mm}
\end{figure}

We evaluate the performance using the Success Rate (SR) metric, where a run is considered successful
if the agent manages to reach the goal-point within $120$~seconds.
Apart from evaluating in the training environment, we set up three realistic, geometrically complex
TartanGround~\cite{patel2025tartanground} meshes in Gazebo simulation for realistic
out-of-distribution evaluation (Fig. \ref{fig:sru_sim_env}).
These consist of Industrial Hangar~(b), Abandoned Cable~(c), and the Modern City Downtown~(d)
environments.
These environments depict realistic scenarios for deployments and are quite different from the
training environment~(a), which primarily consists of mazes, pits, and staircases.
We randomly sample $100$ navigable start-end goal pairs in each of the environments within a range
of \SI{10}{m} to \SI{80}{m}.
We evaluate each policy $3$ times and report the average SR.
We discuss the results obtained in the simulation in the following subsection, and real-world
experiments in Sec.~\ref{sec:sru_real}.

\subsubsection{Results}

The results have been summarized in Tab.~\ref{tab:nav_results}.
While the VAE baseline achieves similar performance to DeFM during training, DeFM demonstrates
improved performance during testing across all three Gazebo environments.
Upon a closer examination of the individual failure cases, we notice that, in general, the policy
utilizing our DeFM model performs better at avoiding thin and out-of-distribution (OOD) obstacles,
such as fences, traffic signs, lamp posts, etc. We hypothesize that DeFM better recognizes these OOD
obstacles and intrinsically assigns higher costs, which the VAE might fail to do. 

\begin{figure}[t]
    \centering
    \includegraphics[width=0.95\linewidth]{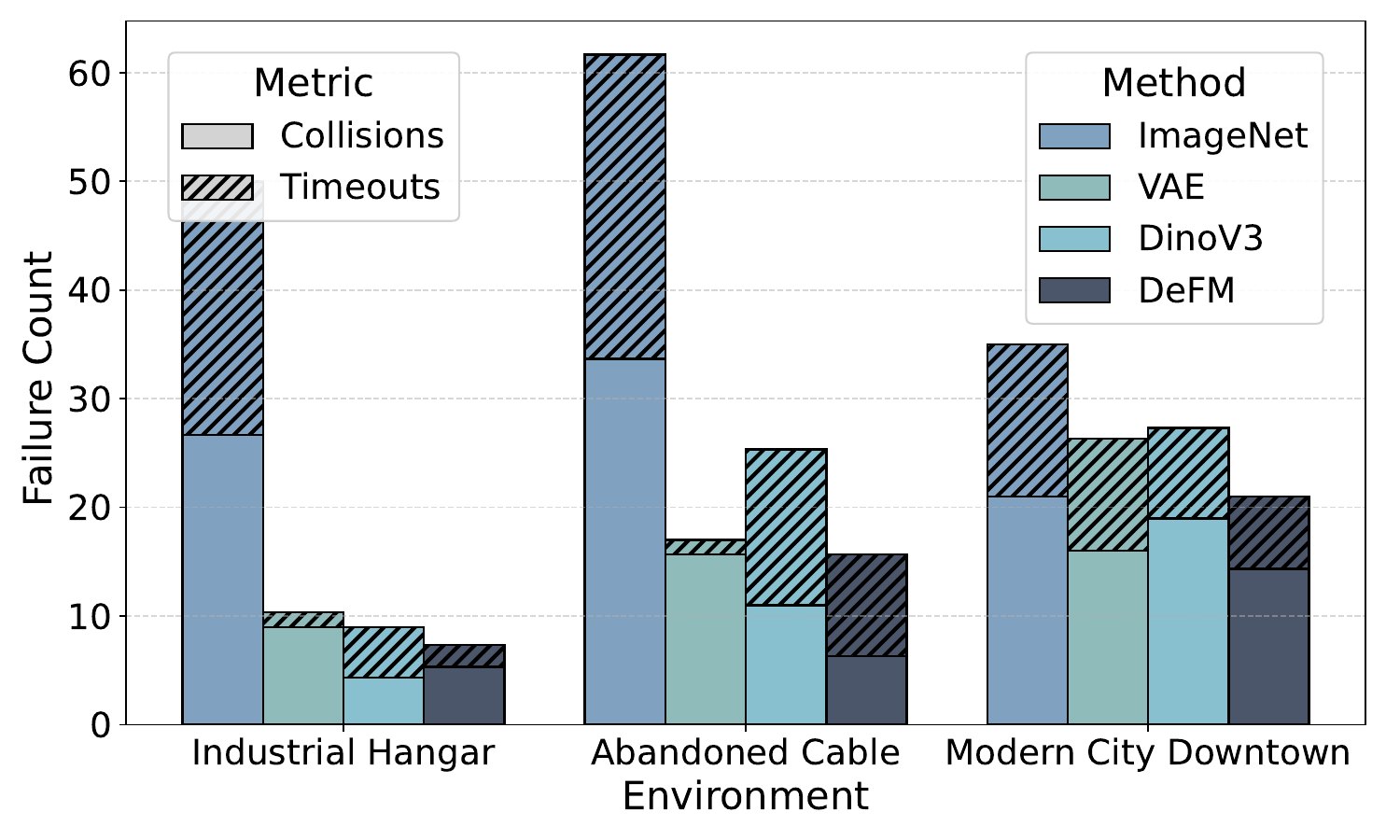}
    \caption{Split of navigation failure count ($\downarrow$) for embodiment aware navigation
    evaluation across the test environments.}
    \label{fig:sru_failure}
    \vspace{-4mm}
\end{figure}

In Fig. \ref{fig:sru_failure}, we quantitatively analyze the failure cases by categorizing them into
collisions and timeouts.
Collision failures generally indicate a perception failure arising primarily due to the encoder,
whereas timeouts typically occur when the navigation policy gets stuck in a local minima.
We notice that DeFM consistently has lower collision failures than the other three baselines,
highlighting the better geometric and semantic understanding of the environment.

Finally, we notice that DINOv3 performs worse than the VAE, however, the largest drop in performance
on the test environments occurs for the ImageNet baseline.
The poor performance of the ImageNet baseline can be attributed to a lack of dense spatial
pretraining.
For DINOv3, we hypothesize that the underperformance occurs from the model's inability to retain
metric depth awareness.
Unlike DeFM, where the input normalization explicitly attempts to preserve and learn across vast
metric scales, DINOv3 treats depth simply as another intensity channel.
This result further demonstrates the importance of metric awareness for successful transfer to
embodiment-aware robotic tasks.

\begin{table}[t]
\centering
\caption{Navigation Success Rate (SR \%) ($\uparrow$) for embodiment aware navigation.}
\label{tab:nav_results}
\renewcommand{\arraystretch}{1.2}
\begin{tabular}{@{}lcccc@{}}
\toprule
{Model} & {Training} & {\begin{tabular}[c]{@{}c@{}}Industrial\\ Hangar\end{tabular}} &
{\begin{tabular}[c]{@{}c@{}}Abandoned\\ Cable\end{tabular}} &
{\begin{tabular}[c]{@{}c@{}}Modern City\\ Downtown\end{tabular}}\\
\midrule
ImageNet & 77.61 & 50.00 & 38.33 & 65.00 \\
VAE & 90.23 & 89.67 & 83.00 & 73.67 \\
DINOv3 & 83.64 & 90.00 & 74.67 & 72.67 \\
DeFM & \textbf{90.31} & \textbf{92.67} & \textbf{84.33} & \textbf{79.00} \\
\bottomrule
\end{tabular}%
\end{table}


\subsubsection{Real World Deployments}
\label{sec:sru_real}
\begin{figure}[t]
    \centering
    \includegraphics[width=\linewidth]{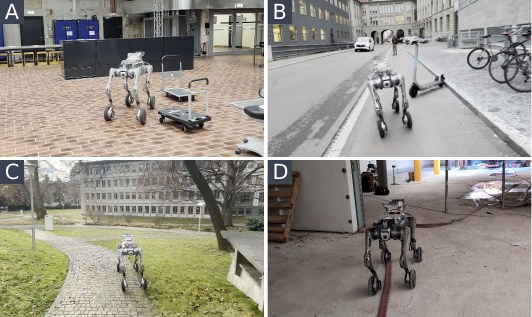}
    \caption{Real world deployments using a Unitree B2W robot in diverse environments- (A) Indoor,
    (B) Urban, (C) Park, (D) Construction Site.}
    \label{fig:sru_real}
    \vspace{-4mm}
\end{figure}

\begin{figure}[ht]
    \centering
    \includegraphics[width=\linewidth]{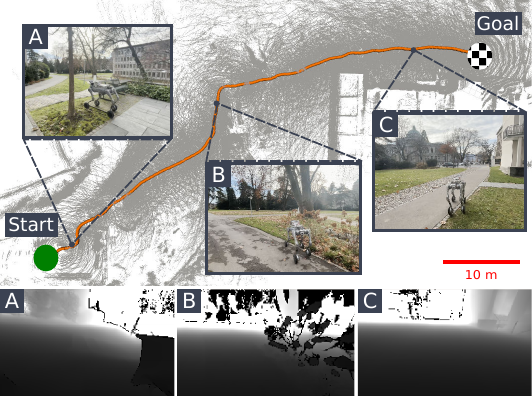}
    \caption{Real world deployment in a park environment with a goal point at around \SI{90}{m}.}
    \label{fig:sru_uni_nav}
    \vspace{-4mm}
\end{figure}

To qualitatively demonstrate successful sim-to-real transfer, we test our DeFM encoder-based policy in four
diverse, challenging real-world environments, including a confined obstacle-rich indoor setting, an
outdoor urban environment, an unstructured park environment, and a high-clutter construction site
(Fig. \ref{fig:sru_real}).
We test multiple start-goal endpoints ranging from \SI{5}{m} to \SI{100}{m} and observe that our
policy generalizes effectively to these environments, handling diverse obstacles such as
pedestrians, scooters, barriers, furniture, etc., as well as navigating uneven terrains.
We visualize one of the deployments from the park environment in Fig.~\ref{fig:sru_uni_nav} and
refer to the supplementary video for more deployments.\par

\subsubsection{Discussions and Limitations}

While the authors in \cite{yang2025improving} also demonstrate sim-to-real transfer in diverse
environments, our result is particularly impressive since, unlike the VAE, which needs to be
pretrained with a specific noise model, depth clippings (\SI{0.1}{m} and \SI{10}{m}), and additional
hand-engineering, our DeFM encoder works out of the box without any task-specific heuristics.

However, one of the failure cases that we observe is that the policy occasionally struggles with
very small or thin obstacles such as wires, poles, or mesh structures (even in simulation).
We attribute this primarily to the low spatial resolution of the depth input ($64\times40$) used
during the policy training, a design choice made by the authors in \cite{yang2025improving} to ease
the sim-to-real transfer and computational constraints.
Despite the low resolution, the policy reliably handles complex clutter, indicating that DeFM
already provides strong features even under heavy compression.
We expect that scaling simulation diversity and input resolution will allow the policy to realize
the full potential of the DeFM features, improving sensitivity to fine-grained geometry and
mitigating these remaining failure cases.





\subsection{Manipulation: Dexterous Grasping}
\label{sec:dex_grasping}
\subsubsection{Setup}

\begin{figure}[t]
    \centering
    \includegraphics[width=0.325\linewidth, trim={350 0 350 100}, clip]{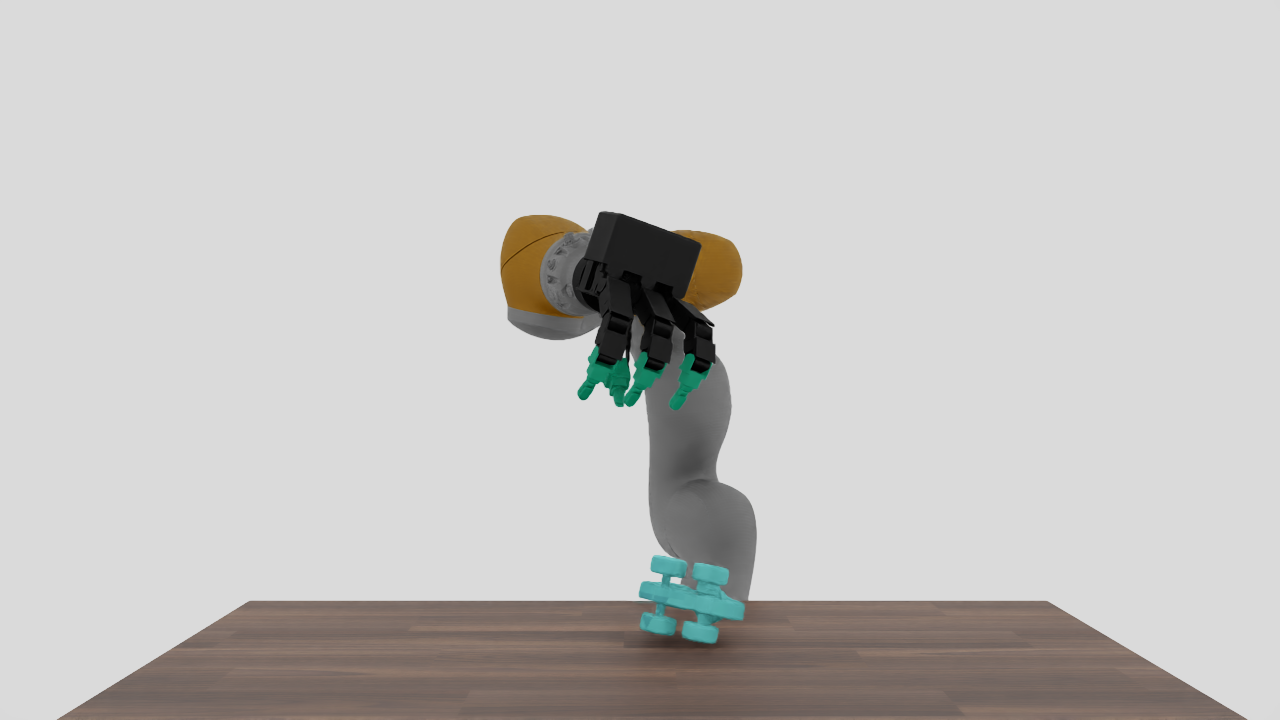}
    \includegraphics[width=0.325\linewidth, trim={350 0 350 100}, clip]{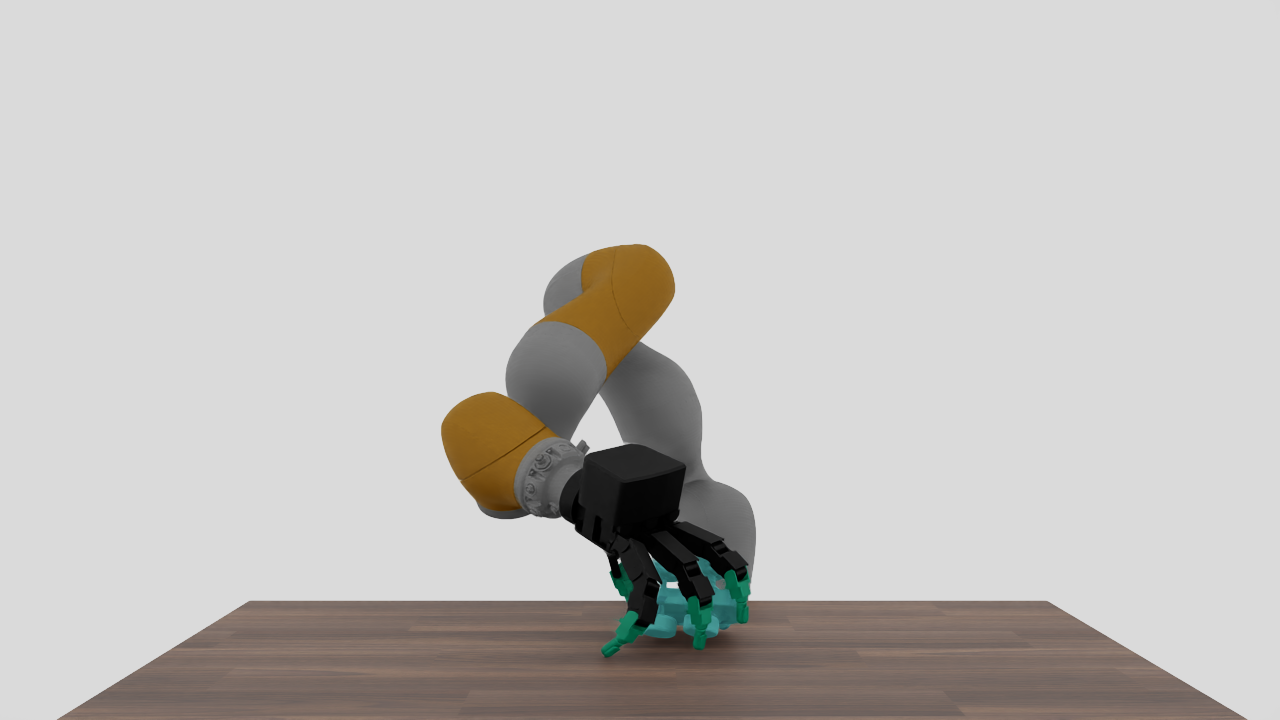}
    \includegraphics[width=0.325\linewidth, trim={350 0 350 100}, clip]{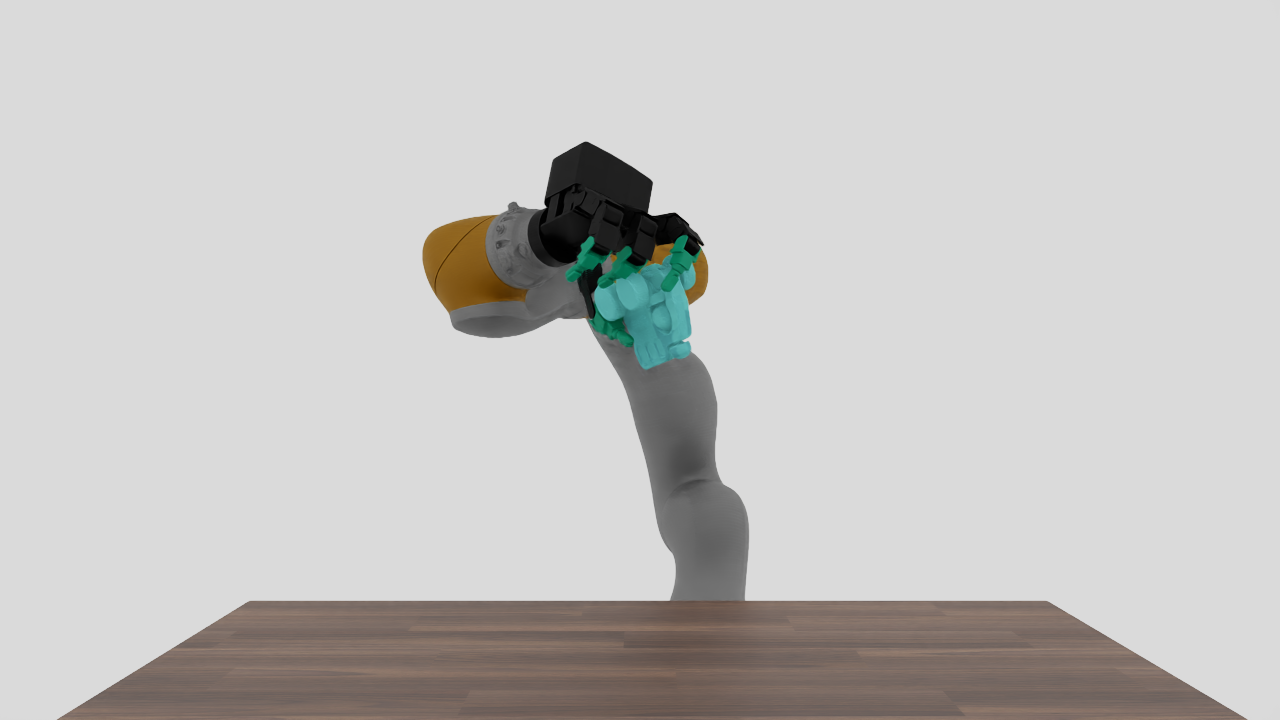}
    \\
    \vspace*{2pt}
    \includegraphics[width=0.325\linewidth]{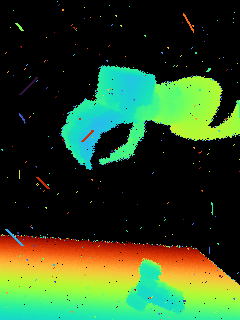}
    \includegraphics[width=0.325\linewidth]{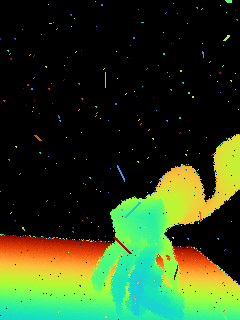}
    \includegraphics[width=0.325\linewidth]{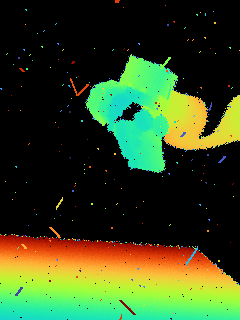} \\
    \caption{Rollout of DeFM model for dexterous grasping using the Kuka-Allegro setup.
    Bottom: The simulated noisy depth images used during training.}
    \vspace{-4mm}
    \label{fig:dex_mani_seq}
\end{figure}

We consider the dexterous arm-hand grasping from DexTRAH~\cite{singh2024dextrah} (Fig.
\ref{fig:dex_mani_seq}).
Following the prior work, the teacher is trained using state-based information, comprising of
privileged information such as the object state and its one-hot encoding.
The teacher policy is distilled into a student policy, which receives depth images instead of stereo
RGB images.
The depth images are encoded using DeFM and concatenated with the proprioceptive information before
feeding it to an LSTM and MLP architecture~\cite{singh2024dextrah}.
We use a ResNet-18 encoder for all baselines.


We compare two training setups: frozen pretrained networks and end-to-end finetuning during
teacher-student policy distillation.
We also compare these with an encoder trained from scratch during the student distillation process.
The training hyperparameters are kept the same as the prior work~\cite{singh2024dextrah}.
The student policies are trained using Isaac Lab~\cite{mittal2025isaac} using distributed training
on eight NVIDIA L40S GPUs, each running 256 parallel environments.
The environment provides depth observations augmented with speckle noise, Gaussian noise, pixel
dropout, and stick noise~\cite{lum2024dextrahg}.

\begin{figure}
    \centering
    \includegraphics[width=\linewidth]{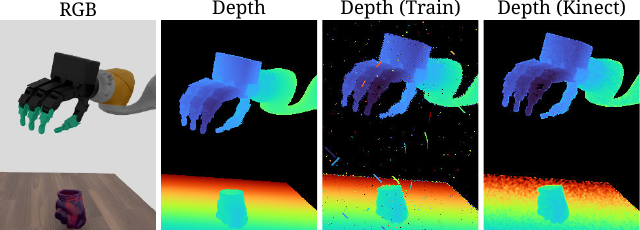}
    \caption{Depth images input to the dexterous grasping policy.
    During training, images are augmented with speckles, dropout, and stick noise.
    For evaluations, we also consider depth images augmented with the Kinect noise
    model~\cite{handa2014kinect}.}
    \label{fig:dextrah_depth_noises}
    \vspace{-2mm}
\end{figure}

\subsubsection{Results}

We evaluate the trained policies in two scenarios: (i) the same noise model as during training, and
(ii) using the Kinect noise model~\cite{handa2014kinect}, which adds edge and distance noise to the
depth image, to evaluate robustness to realistic sensor noise.
A representative scenario with the different noise models is shown in
Fig.~\ref{fig:dextrah_depth_noises}.
Tab.~\ref{tab:dextrah_results} reports the average success rate (normalized to the teacher policy)
across 2,000 episodes.

The fine-tuned DeFM model achieves the highest performance across both scenarios, and fine-tuning
improves results for all models.
Notably, the frozen DeFM encoder outperforms both the scratch-trained encoder and fine-tuned
ImageNet encoder, while achieving competitive performance relative to the fine-tuned DINOv3 and VAE
baselines.
This highlights the strong geometric and semantic priors learnt by DeFM.
\par

Under the alternative Kinect noise model, frozen models experience a significant performance drop,
likely because the MLP on top of frozen features overfits to the training depth images.
Among the frozen models, DeFM provides more consistent features, resulting in a smaller decline.
Fine-tuned DeFM is the least affected by changes in the noise model, demonstrating both the
adaptability and stability of its task-specific features, while finetuned VAE severely overfits to
the training noise model.



\begin{figure}[t]
    \centering
    \includegraphics[width=\linewidth]{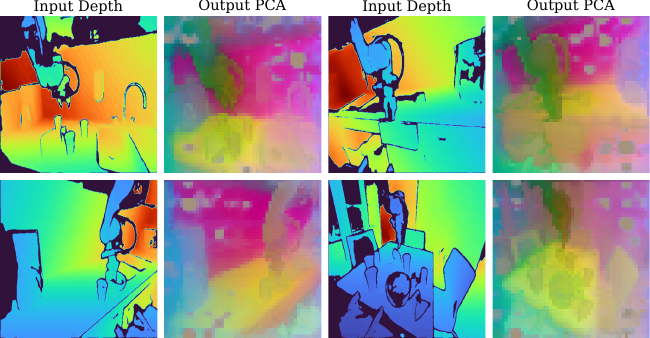}
    \caption{PCA visualization for real-world images from different scenes of the DROID
    dataset~\cite{khazatsky2024droid}.}
    \label{fig:pca_droid}
    \vspace{-5mm}
\end{figure}

Lastly, we present qualitative results by performing PCA on real-world depth images captured with a
ZED camera from the DROID dataset~\cite{khazatsky2024droid}.
In the complex, cluttered kitchen scenes, the DeFM feature space exhibits a clear structure: PCA
embeddings cluster different scene components such as the countertop, background surfaces, robot
arm, and manipulable objects (Fig. \ref{fig:pca_droid}).

\begin{table}[t]
\centering
\caption{Comparison of different model configurations in simulation for dexterous grasping.
Performance is normalized relative to the teacher’s performance and averaged across 2,000 episodes.
All models use ResNet-18 architecture.}
\label{tab:dextrah_results}
\renewcommand{\arraystretch}{1.2}
\setlength{\tabcolsep}{5pt}
\resizebox{\columnwidth}{!}{%
\begin{tabular}{@{}lcccc@{}}
\toprule
\multirow{2}{*}{{Pretraining}} & \multicolumn{2}{c}{\textbf{Frozen}} &
\multicolumn{2}{c}{\textbf{Fine-Tuned}} \\
\cmidrule(lr){2-3} \cmidrule(lr){4-5}
& {Training Noise} & {Kinect Noise} & {Training Noise} & {Kinect Noise} \\
\midrule
Scratch & --- & --- & 0.7774 & 0.7247 \\
ImageNet & 0.6576 & 0.0043 & 0.7950 & 0.7366 \\
VAE & {0.6894} & {0.0219} & {0.8105} & {0.0379} \\
DINOv3 & 0.6532 & 0.2075 & 0.8243 & 0.7837 \\
DeFM & \textbf{0.8089} & \textbf{0.4860} & \textbf{0.8936} & \textbf{0.8763} \\
\bottomrule
\end{tabular}%
}
\vspace{-4mm}
\end{table}

\subsection{Locomotion: Quadruped Ladder Climbing}
\label{sec:ladder_climbing}

\subsubsection{Setup}

\begin{figure}[t]
    \centering
    \includegraphics[width=\linewidth]{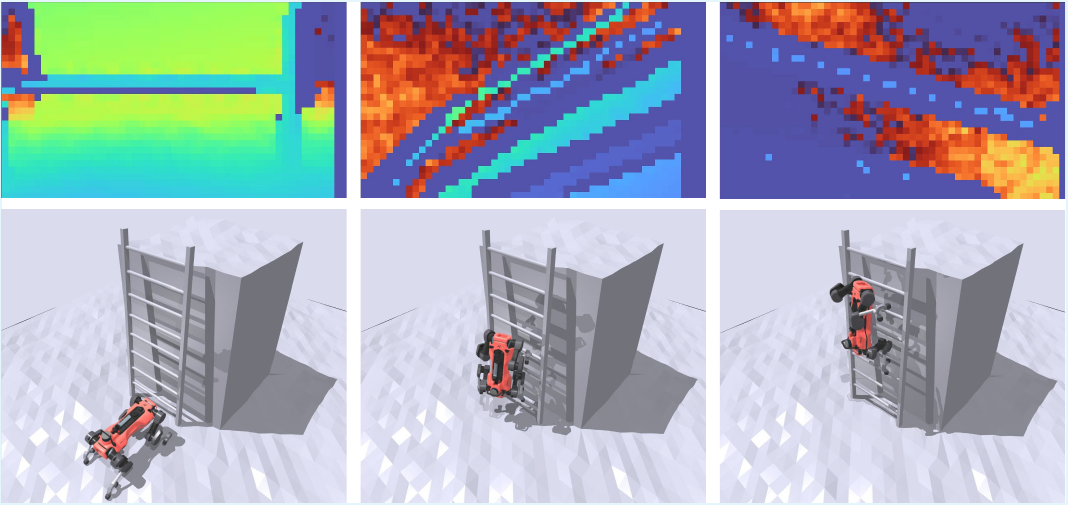}
    \caption{ANYmal climbing a ladder, shown together with noisy depth image (top row) from onboard
    camera.}
    \label{fig:ladder_climbing}
\end{figure}

To evaluate the performance of DeFM on a locomotion task, we consider perceptive quadrupedal ladder
climbing (Fig. \ref{fig:ladder_climbing}) by adapting techniques from~\cite{vogel2024robust}.
The controller is trained using a teacher–student setup where the teacher is trained with privileged
information that the perceptive student learns to mimic.
The student network consists of a CNN encoder, which produces the depth embeddings, which are
concatenated with the proprioceptive measurements and processed by an RNN network to produce the
actions.
For the baselines, we follow \cite{rudin2025parkour} and train a CNN encoder from scratch, and use a
frozen RegNetX-400MF encoder for all other baselines.
The policy is trained in IsaacGym~\cite{makoviychuk2021isaac} using 1,024 parallel environments,
collecting 120 steps per batch over 10,000 training epochs. 



\subsubsection{Results}

\begin{table}[t]
\centering
\caption{Comparison of different models on the Ladder Climbing task in terms of Climb Success Rate}
\label{tab:ladder_results}
\renewcommand{\arraystretch}{1.2}
\setlength{\tabcolsep}{5pt} 
\begin{tabular}{@{}lccccc@{}}
\toprule
{Model} & {CNN-Scratch} & {ImageNet} & {VAE} & {DINOv3} & {DeFM} \\
\midrule
{SR \% ($\uparrow$)} & \textbf{90.45} & 89.29 & 89.75 & 89.59 & 90.14 \\
\bottomrule
\end{tabular}

\end{table}

We evaluate the trained polices across a setup consisting of varying ladder rung radii (15 mm to 45
mm with 5 mm increments) and ladder angles (70\textdegree\ - 90\textdegree\ with 5\textdegree\ 
intervals) while applying random disturbances.
Each combination is tested for 750 episodes with 1024 environments.
We use the Success Rate (SR) metric for comparison, where the episode is considered a success if the
robot manages to climb the ladder within 15 seconds.
As seen in Tab.~\ref{tab:ladder_results}, DeFM performs the best among all frozen models while being
competitive against the CNN baseline trained from scratch.
This demonstrates that our encoder can be transferred to challenging locomotion tasks, without any
task-specific fine-tuning.

\begin{figure}[t]
    \centering
    \includegraphics[width=\linewidth]{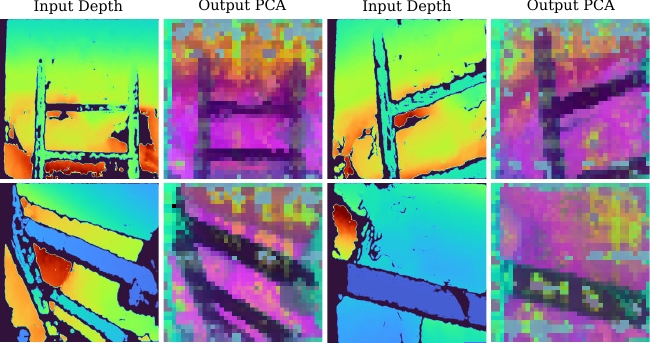}
    \caption{PCA visualization for real-world images collected during deployment of ANYmal
    ladder-climbing in the wild.}
    \label{fig:pca_ladder}
    \vspace{-5mm}
\end{figure}

Since our model was trained to handle sensor noise and varying depth data, we hypothesize an even
stronger generalization for sim-to-real transfer compared to the CNN baseline.
Preliminarily, we present some qualitative results by performing PCA on depth images collected
during a ladder-climbing deployment.
As seen in Fig. \ref{fig:pca_ladder}, our DeFM is able to consistently assign similar features to
the ladder structure despite the heavy noise inherent in real-world stereo cameras.


%% file: chapters/7_limitations.tex
While DeFM demonstrates strong generalization and robust feature extraction across diverse robotic
domains, we acknowledge several limitations that pave the way for future research:
\subsubsection{Dataset Curation and Model Scaling} Although we aggregate numerous open-source depth
datasets, our overall data scale and curation strategies remain limited compared to recent RGB
foundation models like DINOv3.
For instance, our current balancing is relatively naive: we subsample certain sequential datasets
(e.g., GrandTour, ARKitScenes, DROID) at a fixed frequency of 1–5 Hz, while using the entirety of
others (e.g., TartanGround, TartanAir).
Data deduplication is a critical next step.
Moreover, following established scaling laws, we anticipate that coupling these improved curation
strategies with increases in overall dataset size, model capacity, and training iterations will
yield progressively more powerful depth foundation models.
\subsubsection{Reliance on MDE Data} To achieve diversity, a significant portion of our pretraining
corpus relies on Monocular Depth Estimation (MDE).
We acknowledge that this can introduce geometric artifacts, relative scale ambiguity, and inherited
RGB biases into the representations, which real or synthetic metric datasets avoid.
In particular, the MDE depth maps are not calibrated to metric depth but converted to a fixed
pseudo-metric depth range, so the fixed-range channels do not carry physical scale for this portion
of the data.
Using metric MDE models or aligning their predictions to metric scale is a natural extension.
\subsubsection{Metric and Lateral Ambiguity} While our input normalization is metric-aware, absolute
scale ambiguities can still emerge, particularly in the lateral direction due to missing intrinsics.
Future work will explore specific pretraining and architectural modifications, such as integrating
learnable camera intrinsics, explicit metric tokens, or auxiliary depth reconstruction losses, to
strictly enforce metric depth scale.
\subsubsection{Architectural Artifacts} As observed in Figs.~\ref{fig:highlight}-V
and~\ref{fig:pca_ladder}, the learned representations occasionally exhibit patch-level artifacts, a
known limitation of standard ViT architectures.
These anomalies can likely be mitigated in future iterations through architectural refinements, such
as the inclusion of register tokens~\cite{darcet2023vision}.
\subsubsection{Extension to LiDARs} Scaling DeFM to process measurements from solid-state LiDARs and
LiDAR range-image representations would significantly expand its applicability beyond standard
stereo and active IR depth sensors.

%% file: chapters/8_conclusion.tex
In this work, we introduced DeFM, a family of depth image encoders pretrained using self-supervision
on a curated depth dataset of \SI{60}{M} depth images.
DeFM learns robust geometric and semantic features that generalize across tasks, environments, and
sensors.
Our results show that DeFM serves as an effective off-the-shelf depth encoder for a wide range of
robot perception and control tasks, including classification, segmentation, navigation, locomotion,
and manipulation.
Moreover, it enables robust sim-to-real transfer across diverse real-world settings. 

A key outcome of this work is the effectiveness of our distilled variants, which retain most of the
representational strength of the large model while offering substantial gains in efficiency.
These compact models make it practical to deploy strong depth representations on
resource-constrained robotic systems. 
DeFM has the potential to streamline depth perception across a wide range of robotic systems.
A single frozen depth encoder could replace many of the engineering-heavy pipelines still common
today, such as reliance on elevation maps for perceptive locomotion or task-specific depth
preprocessing stages.
Our results show that frozen depth foundation models can enable fast and efficient RL training,
pointing toward a future where depth-based representations offer a simple, general, and
high-performing alternative to task-specific depth-processing modules.





%% file: chapters/9_appendix.tex
Tab.~\ref{tab:app_pretrain} lists the configuration used to pretrain DeFM-L/14, where settings
modified with respect to DINOv2~\cite{oquab2023dinov2} are marked with $\dagger$.
The depth-specific augmentations are given in Tab.~\ref{tab:app_aug}, and the probing protocols in
Tabs.~\ref{tab:app_probe} and~\ref{tab:app_seg}.

\begin{table}[h]
\centering
\caption{Pretraining configuration of DeFM-L/14 ($\dagger$: modified w.r.t.\ DINOv2).}
\label{tab:app_pretrain}
\footnotesize
\setlength{\tabcolsep}{4pt}
\begin{tabular}{@{}ll@{}}
\toprule
Setting & Value \\
\midrule
Backbone & ViT-L/14, stochastic depth 0.4 \\
Objective & DINO + iBOT + KoLeo (weights 1.0 / 1.0 / 0.1) \\
Projection heads & separate DINO and iBOT heads, 131{,}072 prototypes \\
iBOT masking & probability 0.5, mask ratio $\mathcal{U}[0.1, 0.5]$ \\
Centering & Sinkhorn-Knopp \\
Teacher temperature & 0.04 $\rightarrow$ 0.07 \\
Teacher momentum & 0.994 $\rightarrow$ 1.0 (cosine) \\
Optimizer & AdamW ($\beta_1 = 0.9$, $\beta_2 = 0.999$), gradient clip 3.0 \\
Learning rate & $3.5 \times 10^{-4}$, $100k$ warmup, cosine decay to $10^{-6}$ \\
Weight decay & 0.04 $\rightarrow$ 0.2 (cosine) \\
Batch size / iterations & 3{,}072 (96 GPUs) / $625k$ \\
Global crops & $2 \times 224^2$, scale $[0.32, 1.0]$ \\
Local crops & $8 \times 98^2$, scale $[0.05, 0.32]$ \\
Precision & float16 mixed precision, FSDP \\
\midrule
Training data$^\dagger$ & curated depth dataset (Tab.~\ref{tab:datasets}) \\
Input$^\dagger$ & 3-channel log-depth (Sec.~\ref{sec:normalization}) \\
Augmentations$^\dagger$ & depth-specific (Tab.~\ref{tab:app_aug}) \\
\bottomrule
\end{tabular}
\end{table}

\begin{table}[h]
\centering
\caption{Depth-specific augmentations, applied after cropping and horizontal flipping; $p$ denotes
the probability of applying an augmentation to a crop. Color jittering, grayscale conversion, and
solarization of DINOv2 are removed.}
\label{tab:app_aug}
\footnotesize
\setlength{\tabcolsep}{4pt}
\begin{tabular}{@{}lccc@{}}
\toprule
Augmentation & Global 1 & Global 2 & Local \\
\midrule
Intensity jitter$^{a}$, $p$ & 0.8 & 0.8 & 0.8 \\
\midrule
Gaussian blur, $p$ & 1.0 & 0.1 & 0.5 \\
\quad max.\ radius & 2.0 & 1.5 & 1.5 \\
\midrule
Gaussian noise, $p$ & 0.3 & 0.2 & 0.25 \\
\quad std.\ $\sigma$ & 0.03 & 0.02 & 0.018 \\
\midrule
Pixel dropout$^{b}$, $p$ & 0.3 & -- & 0.2 \\
\quad max.\ ratio & 10\% & -- & 5\% \\
\bottomrule
\end{tabular}
\\[2pt]
\raggedright
$^{a}$ Per-channel brightness and contrast, $\pm 20\%$ each.
$^{b}$ Shared mask across channels, simulating missing sensor depth.
\end{table}

\subsubsection*{Classification Probing}
For classification, we follow the linear probing protocol of DINOv2~\cite{oquab2023dinov2}
(Tab.~\ref{tab:app_probe}).

\begin{table}[h]
\centering
\caption{Probing protocol for classification.}
\label{tab:app_probe}
\footnotesize
\setlength{\tabcolsep}{4pt}
\begin{tabular}{@{}ll@{}}
\toprule
Setting & Value \\
\midrule
Features (ViT) & cls token of last 1 or 4 blocks (+ avg.\ patch tokens)$^{a}$ \\
Features (CNN) & pooled feature (+ pooled spatial features)$^{a}$ \\
Input resolution & $224^2$ \\
Augmentations & train: resize 256, random crop, flip; \\
 & test: resize 256, center crop \\
Head & linear \\
Optimizer & SGD, momentum 0.9, no weight decay \\
Learning rate & $10^{-4}$ to $0.1$ (10 values)$^{a}$, cosine decay \\
Batch size / epochs & 256 / 3 \\
Train / test split & 1.28M / 50k images \\
KNN & $k = 20$, temperature 0.07, cosine similarity \\
\bottomrule
\end{tabular}
\\[2pt]
\raggedright
$^{a}$ Grid search over all combinations; the best validation accuracy is reported.
\end{table}

\subsubsection*{Segmentation Probing}
All encoders are evaluated with a frozen linear probe on the five datasets of
Tab.~\ref{tab:seg_datasets}.
Each depth map is resized to a fixed per-dataset resolution (Tab.~\ref{tab:app_seg}), converted to
the encoder's input format (Sec.~\ref{sec:baselines} for the RGB baselines), and center-padded to a
multiple of the patch size.
The last-layer patch tokens of the frozen encoder (for CNN students, the $F'/16$ BiFPN features) are
pre-extracted, so no data augmentation is applied.
A linear layer maps each patch token to class logits, which are bilinearly upsampled to the input
resolution.
The probe is trained with pixel-wise cross-entropy, ignoring unlabeled pixels, using AdamW (weight
decay $10^{-4}$), a one-cycle learning rate schedule with linear warmup and cosine decay, a batch
size of 32, mixed precision, and gradient clipping at 1.0.
For the main results (Tab.~\ref{tab:seg_results}), the peak learning rate is selected from
$\{10^{-3}, 5 \times 10^{-3}\}$ for each encoder and dataset, while the ablations
(Tab.~\ref{tab:ablations}) use a single fixed learning rate; we report the mIoU over all classes.

\begin{table}[h]
\centering
\caption{Dataset-specific settings for seg. probing.}
\label{tab:app_seg}
\footnotesize
\setlength{\tabcolsep}{4pt}
\begin{tabular}{@{}lccc@{}}
\toprule
Dataset & Train / Eval & Resolution ($H \times W$) & Epochs (warmup) \\
\midrule
ScanNet & 19{,}456 / 5{,}436 & $480 \times 640$ & 25 (3) \\
SUN-RGBD & 5{,}285 / 5{,}050 & $530 \times 730$ & 25 (3) \\
TartanGround & 800 / 400 & $640 \times 640$ & 25 (3) \\
OFFSED & 162 / 41 & $360 \times 640$ & 200 (20) \\
GraspNet-1B & 2{,}000 / 600 & $360 \times 640$ & 25 (3) \\
\bottomrule
\end{tabular}
\end{table}

%% file: chapters/acknowledgements.tex
\section*{ACKNOWLEDGMENT}
The authors would like to thank Marco Trentini, Per Frivik, and Linus Kramer for support during the
navigation experiments, and Ankur Handa for insightful discussions and assistance in setting up the
dexterous grasping experiments.
We further thank Ren\'{e} Zurbr\"{u}gg and Arjun Bharadwaj for helpful discussions regarding the
manipulation tasks. 

This work was supported as part of the Swiss AI Initiative by a grant from the Swiss National
Supercomputing Centre (CSCS) under project ID a144 on Alps.
This work was also supported by the Luxembourg National Research Fund (Ref. 18990533), and the Swiss
National Science Foundation (SNSF) through projects No.200021E\_229503 and No.227617.

The author(s) used generative AI to assist with drafting and editing the manuscript.
All content was reviewed by the authors, who take full responsibility for it.


%% file: chapters/10_biography.tex
\makeatletter\def\@IEEEBIOskipN{2\baselineskip}\makeatother 
\newcommand{\biophotoplaceholder}{\fcolorbox{gray}{gray!15}{\parbox[c][1.15in][c]{0.88in}{\centering\scriptsize
\textcolor{red}{photo}}}}

\begin{IEEEbiography}[{\includegraphics[width=1in,height=1.25in,clip,keepaspectratio]{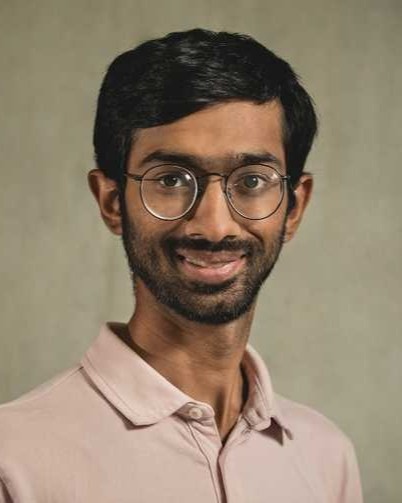}}]
{Manthan Patel}~(Student Member, IEEE)~is a PhD candidate at the Robotic Systems Lab at ETH Zurich
working on foundation models for robot perception and navigation.
He received his M.Sc. in Robotics, Systems \& Control in 2024 from ETH Zurich, prior to which he
received a BS in Mechanical Engineering from IIT Kharagpur in 2021.
His research interests lie in various aspects of field robotics with a particular focus on
large-scale model training and datasets for perception and navigation.
\end{IEEEbiography}

\begin{IEEEbiography}[{\includegraphics[width=1in,height=1.25in,clip,keepaspectratio]{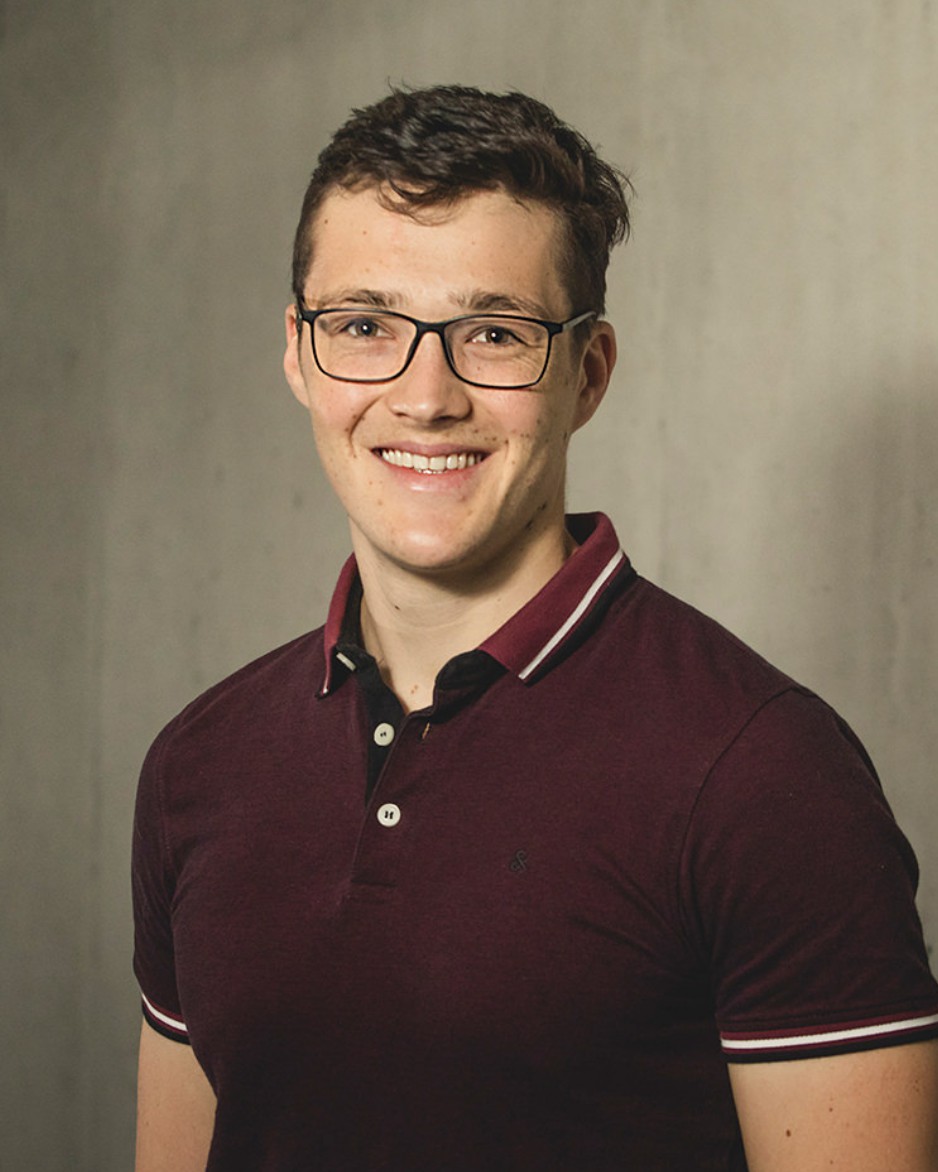}}]
{Jonas Frey}~(Member, IEEE)~is a Postdoctoral Researcher at UC Berkeley and an Affiliate Assistant
Professor at the University of Washington.
He received his Ph.D. from the Robotic Systems Lab at ETH Zurich and the Max Planck Institute for
Intelligent Systems through the MPI ETH Center for Learning Systems.
He earned his M.Sc. in Robotics, Systems \& Control in 2021 from ETH Zurich.
His research interests focus on perception, navigation, and locomotion to enable the robust
deployment of mobile robotic systems.
\end{IEEEbiography}

\begin{IEEEbiography}[{\includegraphics[width=1in,height=1.25in,clip,keepaspectratio]{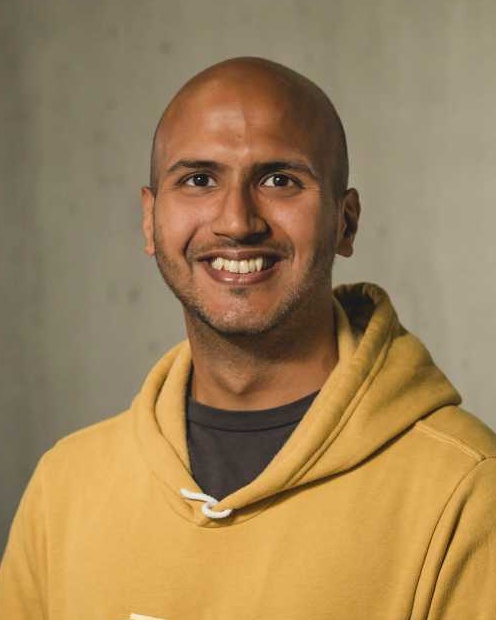}}]
{Mayank Mittal}~is a Senior Research Scientist at NVIDIA, and a Ph.D. student at ETH Zurich under the
supervision of Prof. Marco Hutter.
He received his B.Tech. degree in Electrical Engineering from IIT Kanpur, India, in 2018, and his M.Sc.
degree in Robotics, Systems and Control from ETH Zurich in 2021.
His research interests include reinforcement learning, robot simulation, and legged mobile manipulation,
with an emphasis on contact-rich interactions in real-world environments.
\end{IEEEbiography}

\begin{IEEEbiography}[{\includegraphics[width=1in,height=1.25in,clip,keepaspectratio]{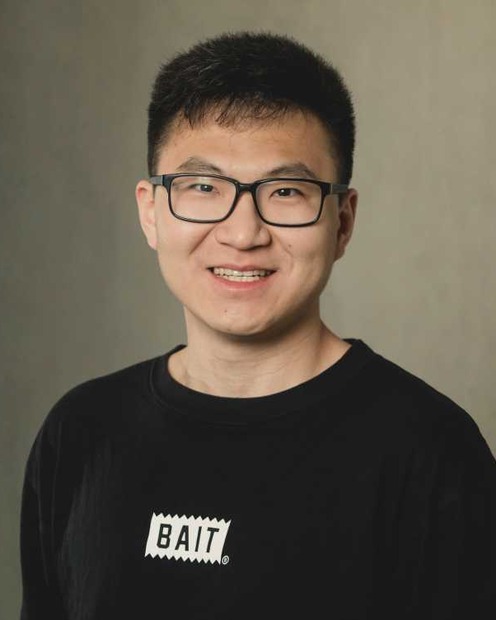}}]
{Fan Yang}~is a Ph.D. researcher at ETH Zurich's Robotic Systems Lab, supervised by Prof. Marco
Hutter and co-supervised by Prof. Sebastian Scherer.
His research focuses on autonomous navigation, perception-driven reinforcement learning, long-horizon
decision-making, and planning in complex environments.
He received his M.S. in Robotic Systems Development from Carnegie Mellon University in 2020,
supervised by Prof. Sebastian Scherer and Prof. Ji Zhang, and his Bachelor of Engineering from
Beihang University.
His work on FAR Planner received the ABB Best Student Paper Award at IROS 2022.
\end{IEEEbiography}

\begin{IEEEbiography}[{\includegraphics[width=1in,height=1.25in,clip,keepaspectratio]{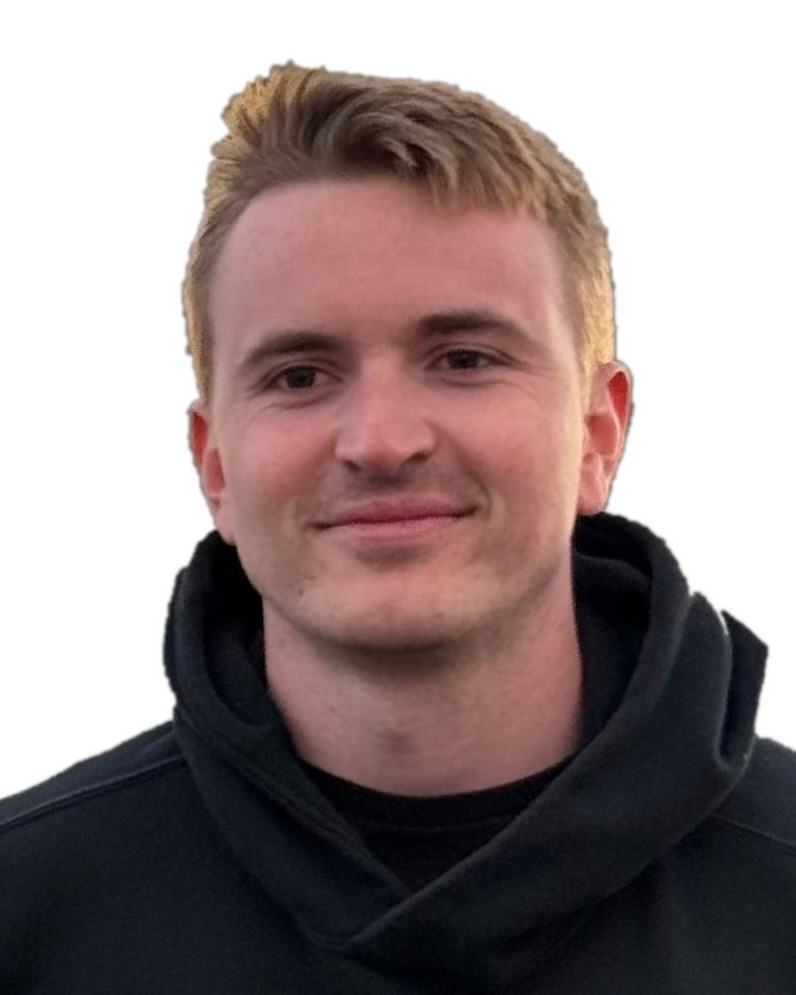}}]
{Alexander Hansson}~is a PhD candidate at the Mobile Robotics Lab at ETH Zurich, where he works on
perceptive locomanipulation for humanoid robots.
He received both his B.Sc. in Mechanical Engineering and his M.Sc. in Robotics, Systems and Control
(2025) from ETH Zurich.
His research interests lie at the intersection of robot learning, perception and whole-body control,
with the goal of enabling legged robots to operate robustly in complex, unstructured environments.
\end{IEEEbiography}

\begin{IEEEbiography}[{\includegraphics[width=1in,height=1.25in,clip,keepaspectratio]{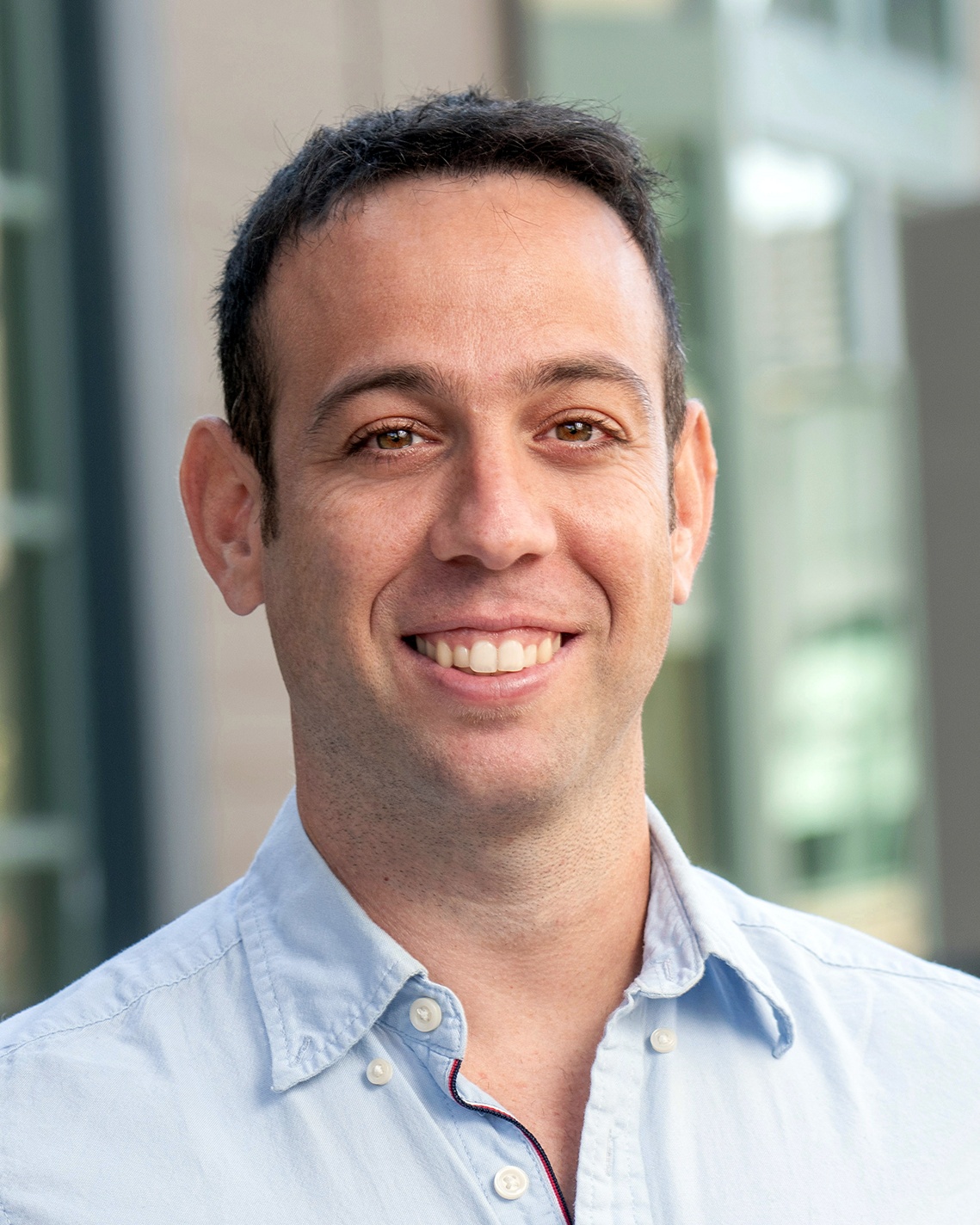}}]
{Amir Bar}~is an Assistant Professor at Imperial College London and a founding member of AMI Labs.
Previously, he was a Research Scientist and Postdoctoral Researcher at FAIR after completing his PhD
at Tel Aviv University.
Amir is a recipient of the Blavatnik PhD Award, a CVPR 2025 Best Paper Honorable Mention Award, and
an ICLR 2026 World Models Workshop Outstanding Paper Award.
His work on video models also won the Ego4D CVPR 2022 PNR Challenge.
Before beginning his PhD, Amir led the AI team at Zebra Medical Vision, which was later acquired, and
developed multiple FDA-approved algorithms that are currently used in clinical practice worldwide.
\end{IEEEbiography}

\begin{IEEEbiography}[{\includegraphics[width=1in,height=1.25in,clip,keepaspectratio]{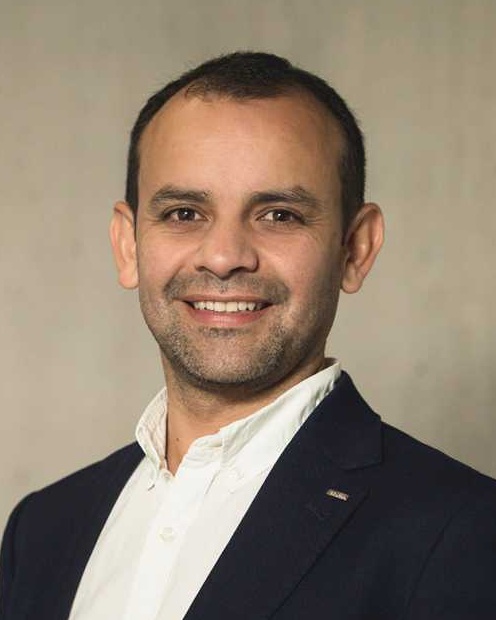}}]
{Cesar Cadena}~(Member, IEEE)~received his PhD (2011) in computer science from the University of
Zaragoza, Spain.
He is a Senior Scientist at ETH Zurich.
His research focuses on equipping robots to understand and interact with an ever-changing, open
world.
Building on his work on simultaneous localization and mapping, place recognition, and semantic and
persistent mapping, he develops robotic scene understanding that goes beyond geometry and semantics
to the physical properties that matter for action.
His current interests include self-supervised learning of terrain properties from the robot's own
experience, foundation models for robotics and open-world spatial AI, reliable perception under novel
conditions, and physically perceptive policies for legged locomotion and mobile manipulation.
\end{IEEEbiography}

\begin{IEEEbiography}[{\includegraphics[width=1in,height=1.25in,clip,keepaspectratio]{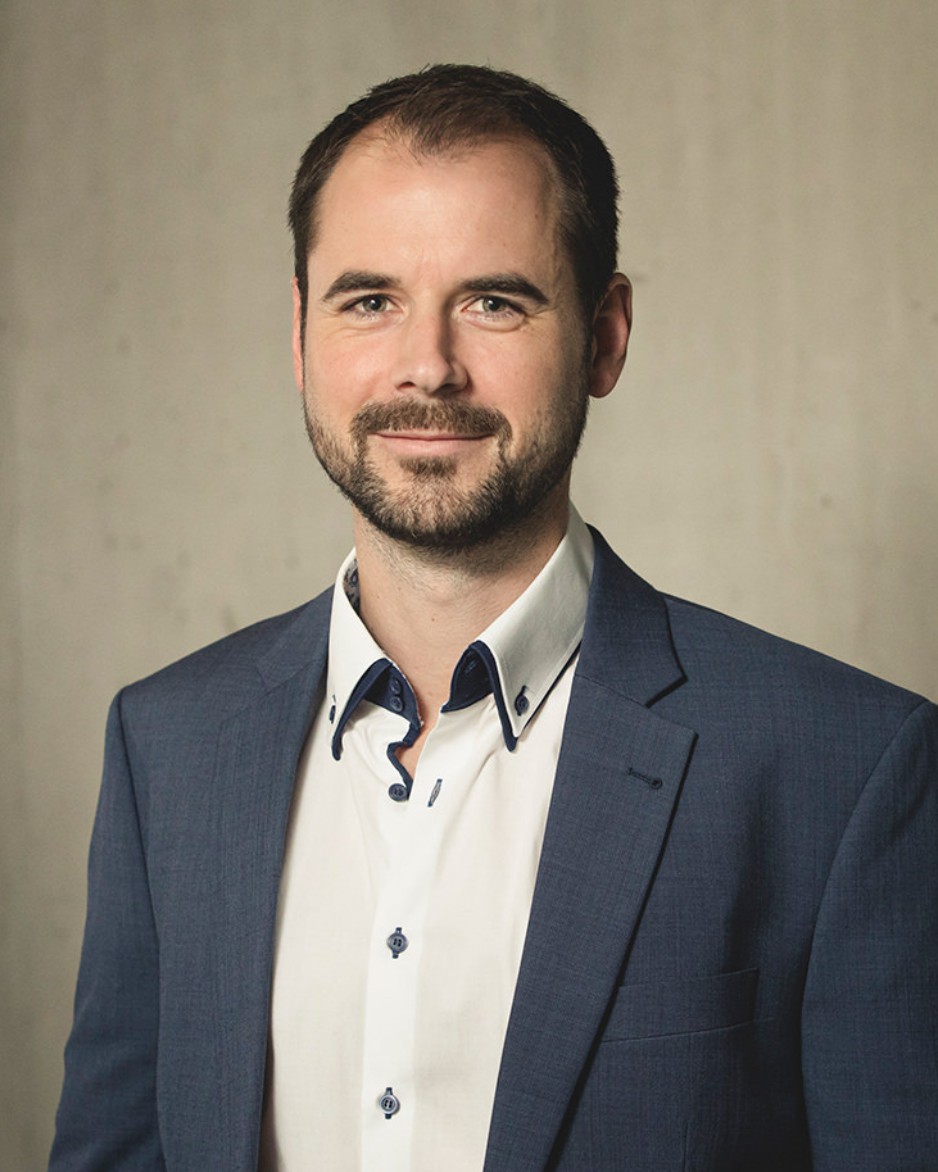}}]
{Marco Hutter}~(Member, IEEE)~received the M.Sc. and Ph.D. degrees in design, actuation, and control
of legged robots from ETH Z\"urich, Z\"urich, Switzerland, in 2009 and 2013, respectively.
He is currently a Professor of robotic systems and the Director of the Center for Robotics,
ETH Z\"urich, and a Co-Founder of several ETH startups, such as ANYbotics and Gravis Robotics.
His research interests include novel machines and actuation concepts, along with
the underlying control, planning, and machine learning algorithms for locomotion and manipulation.
Dr. Hutter was a recipient of the ERC Starting Grant and the winner of the DARPA SubT Challenge.
\end{IEEEbiography}

%% file: root.bbl
\begin{thebibliography}{100}
\providecommand{\url}[1]{#1}
\csname url@samestyle\endcsname
\providecommand{\newblock}{\relax}
\providecommand{\bibinfo}[2]{#2}
\providecommand{\BIBentrySTDinterwordspacing}{\spaceskip=0pt\relax}
\providecommand{\BIBentryALTinterwordstretchfactor}{4}
\providecommand{\BIBentryALTinterwordspacing}{\spaceskip=\fontdimen2\font plus
\BIBentryALTinterwordstretchfactor\fontdimen3\font minus
  \fontdimen4\font\relax}
\providecommand{\BIBforeignlanguage}[2]{{%
\expandafter\ifx\csname l@#1\endcsname\relax
\typeout{** WARNING: IEEEtran.bst: No hyphenation pattern has been}%
\typeout{** loaded for the language `#1'. Using the pattern for}%
\typeout{** the default language instead.}%
\else
\language=\csname l@#1\endcsname
\fi
#2}}
\providecommand{\BIBdecl}{\relax}
\BIBdecl

\bibitem{caron2021emerging}
M.~Caron, H.~Touvron, I.~Misra \emph{et~al.}, ``Emerging properties in
  self-supervised vision transformers,'' in \emph{Proc. IEEE/CVF Int. Conf.
  Comput. Vis. (ICCV)}, 2021, pp. 9630--9640.

\bibitem{oquab2023dinov2}
M.~Oquab, T.~Darcet, T.~Moutakanni \emph{et~al.}, ``{DINOv2}: Learning robust
  visual features without supervision,'' \emph{Trans. Mach. Learn. Res.}, 2024.

\bibitem{simeoni2025dinov3}
O.~Sim{\'e}oni, H.~V. Vo, M.~Seitzer \emph{et~al.}, ``{DINOv3},'' \emph{Trans.
  Mach. Learn. Res.}, 2026.

\bibitem{radford2021learning}
A.~Radford, J.~W. Kim, C.~Hallacy \emph{et~al.}, ``Learning transferable visual
  models from natural language supervision,'' in \emph{Proc. Int. Conf. Mach.
  Learn. (ICML)}, ser. PMLR, vol. 139, 2021, pp. 8748--8763.

\bibitem{kim24openvla}
M.~J. Kim, K.~Pertsch, S.~Karamcheti \emph{et~al.}, ``{OpenVLA}: An open-source
  vision-language-action model,'' in \emph{Proc. Conf. Robot Learn. (CoRL)},
  ser. PMLR, vol. 270, 2025, pp. 2679--2713.

\bibitem{intelligence2025pi_}
K.~Black, N.~Brown, J.~Darpinian \emph{et~al.}, ``$\pi_{0.5}$: A
  vision-language-action model with open-world generalization,'' in \emph{Proc.
  Conf. Robot Learn. (CoRL)}, ser. PMLR, vol. 305, 2025, pp. 17--40.

\bibitem{zhang2024uni}
J.~Zhang, K.~Wang, S.~Wang \emph{et~al.}, ``{Uni-NaVid}: A video-based
  vision-language-action model for unifying embodied navigation tasks,'' in
  \emph{Proc. Robot.: Sci. Syst. (RSS)}, 2025.

\bibitem{shang2024theia}
J.~Shang, K.~Schmeckpeper, B.~B. May \emph{et~al.}, ``{Theia}: Distilling
  diverse vision foundation models for robot learning,'' in \emph{Proc. Conf.
  Robot Learn. (CoRL)}, ser. PMLR, vol. 270, 2025, pp. 724--748.

\bibitem{Nair2022R3MAU}
S.~Nair, A.~Rajeswaran, V.~Kumar, C.~Finn, and A.~Gupta, ``{R3M}: A universal
  visual representation for robot manipulation,'' in \emph{Proc. Conf. Robot
  Learn. (CoRL)}, ser. PMLR, vol. 205, 2023, pp. 892--909.

\bibitem{gavryushin2025maple}
A.~Gavryushin, X.~Wang, R.~J.~S. Malate \emph{et~al.}, ``{MAPLE}: Encoding
  dexterous robotic manipulation priors learned from egocentric videos,''
  \emph{arXiv preprint arXiv:2504.06084}, 2025.

\bibitem{xiao2022masked}
T.~Xiao, I.~Radosavovic, T.~Darrell, and J.~Malik, ``Masked visual pre-training
  for motor control,'' \emph{arXiv preprint arXiv:2203.06173}, 2022.

\bibitem{muratore2022robot}
F.~Muratore, F.~Ramos, G.~Turk, W.~Yu, M.~Gienger, and J.~Peters, ``Robot
  learning from randomized simulations: A review,'' \emph{Front. Robot. AI},
  vol.~9, p. 799893, 2022.

\bibitem{miki2022learning}
T.~Miki, J.~Lee, J.~Hwangbo, L.~Wellhausen, V.~Koltun, and M.~Hutter,
  ``Learning robust perceptive locomotion for quadrupedal robots in the wild,''
  \emph{Sci. Robot.}, vol.~7, no.~62, p. eabk2822, 2022.

\bibitem{rudin2022learning}
N.~Rudin, D.~Hoeller, P.~Reist, and M.~Hutter, ``Learning to walk in minutes
  using massively parallel deep reinforcement learning,'' in \emph{Proc. Conf.
  Robot Learn. (CoRL)}, ser. PMLR, vol. 164, 2022, pp. 91--100.

\bibitem{agarwal2023legged}
A.~Agarwal, A.~Kumar, J.~Malik, and D.~Pathak, ``Legged locomotion in
  challenging terrains using egocentric vision,'' in \emph{Proc. Conf. Robot
  Learn. (CoRL)}, ser. PMLR, vol. 205, 2023, pp. 403--415.

\bibitem{Wijmans2019DDPPOLN}
E.~Wijmans, A.~Kadian, A.~S. Morcos \emph{et~al.}, ``{DD-PPO}: Learning
  near-perfect {PointGoal} navigators from 2.5 billion frames,'' in \emph{Proc.
  Int. Conf. Learn. Represent. (ICLR)}, 2020.

\bibitem{lee2024learning}
J.~Lee, M.~Bjelonic, A.~Reske, L.~Wellhausen, T.~Miki, and M.~Hutter,
  ``Learning robust autonomous navigation and locomotion for wheeled-legged
  robots,'' \emph{Sci. Robot.}, vol.~9, no.~89, p. eadi9641, 2024.

\bibitem{yang2025improving}
F.~Yang, P.~Frivik, D.~Hoeller, C.~Wang, C.~Cadena, and M.~Hutter,
  ``Spatially-enhanced recurrent memory for long-range mapless navigation via
  end-to-end reinforcement learning,'' \emph{Int. J. Robot. Res.}, vol.~45,
  no.~11, pp. 1682--1706, 2026.

\bibitem{lum2024dextrahg}
T.~G.~W. Lum, M.~Matak, V.~Makoviychuk \emph{et~al.}, ``{DextrAH-G}:
  Pixels-to-action dexterous arm-hand grasping with geometric fabrics,'' in
  \emph{Proc. Conf. Robot Learn. (CoRL)}, ser. PMLR, vol. 270, 2025, pp.
  3182--3211.

\bibitem{singh2025end}
R.~Singh, K.~{Van Wyk}, P.~Abbeel, J.~Malik, N.~Ratliff, and A.~Handa,
  ``End-to-end {RL} improves dexterous grasping policies,'' \emph{arXiv
  preprint arXiv:2509.16434}, 2025.

\bibitem{zhang2025robustdexgrasp}
H.~Zhang, Z.~Wu, L.~Huang, S.~Christen, and J.~Song, ``Robust dexterous
  grasping of general objects,'' in \emph{Proc. Conf. Robot Learn. (CoRL)},
  ser. PMLR, vol. 305, 2025, pp. 3035--3050.

\bibitem{hu2019acnet}
X.~Hu, K.~Yang, L.~Fei, and K.~Wang, ``{ACNet}: Attention based network to
  exploit complementary features for {RGBD} semantic segmentation,'' in
  \emph{Proc. IEEE Int. Conf. Image Process. (ICIP)}, 2019, pp. 1440--1444.

\bibitem{dosovitskiy2020image}
A.~Dosovitskiy, L.~Beyer, A.~Kolesnikov \emph{et~al.}, ``An image is worth
  16x16 words: Transformers for image recognition at scale,'' in \emph{Proc.
  Int. Conf. Learn. Represent. (ICLR)}, 2021.

\bibitem{balestriero2023cookbook}
R.~Balestriero, M.~Ibrahim, V.~Sobal \emph{et~al.}, ``A cookbook of
  self-supervised learning,'' \emph{arXiv preprint arXiv:2304.12210}, 2023.

\bibitem{chen2020simple}
T.~Chen, S.~Kornblith, M.~Norouzi, and G.~Hinton, ``A simple framework for
  contrastive learning of visual representations,'' in \emph{Proc. Int. Conf.
  Mach. Learn. (ICML)}, ser. PMLR, vol. 119, 2020, pp. 1597--1607.

\bibitem{he2020momentum}
K.~He, H.~Fan, Y.~Wu, S.~Xie, and R.~Girshick, ``Momentum contrast for
  unsupervised visual representation learning,'' in \emph{Proc. IEEE/CVF Conf.
  Comput. Vis. Pattern Recognit. (CVPR)}, 2020, pp. 9726--9735.

\bibitem{grill2020bootstrap}
J.-B. Grill, F.~Strub, F.~Altch{\'e} \emph{et~al.}, ``Bootstrap your own latent
  -- a new approach to self-supervised learning,'' in \emph{Proc. Adv. Neural
  Inf. Process. Syst. (NeurIPS)}, vol.~33, 2020, pp. 21\,271--21\,284.

\bibitem{he2022masked}
K.~He, X.~Chen, S.~Xie, Y.~Li, P.~Doll{\'a}r, and R.~Girshick, ``Masked
  autoencoders are scalable vision learners,'' in \emph{Proc. IEEE/CVF Conf.
  Comput. Vis. Pattern Recognit. (CVPR)}, 2022, pp. 15\,979--15\,988.

\bibitem{zhou2021ibot}
J.~Zhou, C.~Wei, H.~Wang \emph{et~al.}, ``{iBOT}: Image {BERT} pre-training
  with online tokenizer,'' in \emph{Proc. Int. Conf. Learn. Represent. (ICLR)},
  2022.

\bibitem{assran2023self}
M.~Assran, Q.~Duval, I.~Misra \emph{et~al.}, ``Self-supervised learning from
  images with a joint-embedding predictive architecture,'' in \emph{Proc.
  IEEE/CVF Conf. Comput. Vis. Pattern Recognit. (CVPR)}, 2023, pp.
  15\,619--15\,629.

\bibitem{bar2024stochastic}
A.~Bar, F.~Bordes, A.~Shocher \emph{et~al.}, ``Stochastic positional embeddings
  improve masked image modeling,'' in \emph{Proc. Int. Conf. Mach. Learn.
  (ICML)}, ser. PMLR, vol. 235, 2024, pp. 2944--2958.

\bibitem{zhai2023sigmoid}
X.~Zhai, B.~Mustafa, A.~Kolesnikov, and L.~Beyer, ``Sigmoid loss for language
  image pre-training,'' in \emph{Proc. IEEE/CVF Int. Conf. Comput. Vis.
  (ICCV)}, 2023, pp. 11\,941--11\,952.

\bibitem{jia2021scaling}
C.~Jia, Y.~Yang, Y.~Xia \emph{et~al.}, ``Scaling up visual and vision-language
  representation learning with noisy text supervision,'' in \emph{Proc. Int.
  Conf. Mach. Learn. (ICML)}, ser. PMLR, vol. 139, 2021, pp. 4904--4916.

\bibitem{el2024probing}
M.~El~Banani, A.~Raj, K.-K. Maninis \emph{et~al.}, ``Probing the {3D} awareness
  of visual foundation models,'' in \emph{Proc. IEEE/CVF Conf. Comput. Vis.
  Pattern Recognit. (CVPR)}, 2024, pp. 21\,795--21\,806.

\bibitem{wang2024dust3r}
S.~Wang, V.~Leroy, Y.~Cabon, B.~Chidlovskii, and J.~Revaud, ``{DUSt3R}:
  Geometric {3D} vision made easy,'' in \emph{Proc. IEEE/CVF Conf. Comput. Vis.
  Pattern Recognit. (CVPR)}, 2024, pp. 20\,697--20\,709.

\bibitem{wang2025vggt}
J.~Wang, M.~Chen, N.~Karaev, A.~Vedaldi, C.~Rupprecht, and D.~Novotny,
  ``{VGGT}: Visual geometry grounded transformer,'' in \emph{Proc. IEEE/CVF
  Conf. Comput. Vis. Pattern Recognit. (CVPR)}, 2025, pp. 5294--5306.

\bibitem{lin2025depth}
H.~Lin, S.~Chen, J.~H. Liew \emph{et~al.}, ``Depth anything 3: Recovering the
  visual space from any views,'' in \emph{Proc. Int. Conf. Learn. Represent.
  (ICLR)}, 2026, pp. 141\,261--141\,285.

\bibitem{Ranzinger_2024_CVPR}
M.~Ranzinger, G.~Heinrich, J.~Kautz, and P.~Molchanov, ``{AM-RADIO}:
  Agglomerative vision foundation model reduce all domains into one,'' in
  \emph{Proc. IEEE/CVF Conf. Comput. Vis. Pattern Recognit. (CVPR)}, 2024, pp.
  12\,490--12\,500.

\bibitem{heinrich2025radiov25improvedbaselinesagglomerative}
G.~Heinrich, M.~Ranzinger, H.~Yin \emph{et~al.}, ``{RADIOv2.5}: Improved
  baselines for agglomerative vision foundation models,'' in \emph{Proc.
  IEEE/CVF Conf. Comput. Vis. Pattern Recognit. (CVPR)}, 2025, pp.
  22\,487--22\,497.

\bibitem{kirillov2023segment}
A.~Kirillov, E.~Mintun, N.~Ravi \emph{et~al.}, ``Segment anything,'' in
  \emph{Proc. IEEE/CVF Int. Conf. Comput. Vis. (ICCV)}, 2023, pp. 3992--4003.

\bibitem{yin2024dformer}
B.~Yin, X.~Zhang, Z.-Y. Li, L.~Liu, M.-M. Cheng, and Q.~Hou, ``{DFormer}:
  Rethinking {RGBD} representation learning for semantic segmentation,'' in
  \emph{Proc. Int. Conf. Learn. Represent. (ICLR)}, 2024.

\bibitem{yin2026dformer++}
B.-W. Yin, J.-L. Cao, D.~Xu, M.-M. Cheng, and Q.~Hou, ``{DFormer++}: Improving
  {RGBD} representation learning for semantic segmentation,'' \emph{IEEE Trans.
  Pattern Anal. Mach. Intell.}, vol.~48, no.~7, pp. 7195--7208, 2026.

\bibitem{du2024asymformer}
S.~Du, W.~Wang, R.~Guo, R.~Wang, and S.~Tang, ``{AsymFormer}: Asymmetrical
  cross-modal representation learning for mobile platform real-time {RGB-D}
  semantic segmentation,'' in \emph{Proc. IEEE/CVF Conf. Comput. Vis. Pattern
  Recognit. Workshops (CVPRW)}, 2024, pp. 7608--7615.

\bibitem{zhou2022decoupling}
B.~Zhou, P.~Wang, J.~Wan \emph{et~al.}, ``Decoupling and recoupling
  spatiotemporal representation for {RGB-D}-based motion recognition,'' in
  \emph{Proc. IEEE/CVF Conf. Comput. Vis. Pattern Recognit. (CVPR)}, 2022, pp.
  20\,122--20\,131.

\bibitem{xu2017multi}
X.~Xu, Y.~Li, G.~Wu, and J.~Luo, ``Multi-modal deep feature learning for
  {RGB-D} object detection,'' \emph{Pattern Recognit.}, vol.~72, pp. 300--313,
  2017.

\bibitem{qin2023rgb}
R.~Qin, H.~Ma, B.~Gao, and D.~Huang, ``{RGB-D} grasp detection via depth guided
  learning with cross-modal attention,'' in \emph{Proc. IEEE Int. Conf. Robot.
  Autom. (ICRA)}, 2023, pp. 8003--8009.

\bibitem{tan2026masked}
B.~Tan, C.~Sun, X.~Qin \emph{et~al.}, ``Masked depth modeling for spatial
  perception,'' in \emph{Proc. Eur. Conf. Comput. Vis. (ECCV)}, ser.
  LNCS.\hskip 1em plus 0.5em minus 0.4em\relax Springer, 2026, pp. 453--470.

\bibitem{wu2023masked}
X.~Wu, X.~Wen, X.~Liu, and H.~Zhao, ``Masked scene contrast: A scalable
  framework for unsupervised {3D} representation learning,'' in \emph{Proc.
  IEEE/CVF Conf. Comput. Vis. Pattern Recognit. (CVPR)}, 2023, pp. 9415--9424.

\bibitem{wu2025sonata}
X.~Wu, D.~DeTone, D.~Frost \emph{et~al.}, ``Sonata: Self-supervised learning of
  reliable point representations,'' in \emph{Proc. IEEE/CVF Conf. Comput. Vis.
  Pattern Recognit. (CVPR)}, 2025, pp. 22\,193--22\,204.

\bibitem{zhang2026concerto}
Y.~Zhang, X.~Wu, Y.~Lao \emph{et~al.}, ``Concerto: Joint {2D-3D}
  self-supervised learning emerges spatial representations,'' in \emph{Proc.
  Adv. Neural Inf. Process. Syst. (NeurIPS)}, vol.~38, 2025, pp.
  69\,498--69\,522.

\bibitem{zhang2026utonia}
Y.~Zhang, X.~Wu, Y.~Yang \emph{et~al.}, ``Utonia: Toward one encoder for all
  point clouds,'' in \emph{Proc. Int. Conf. Mach. Learn. (ICML)}, ser. PMLR,
  vol. 306, 2026, pp. 159\,370--159\,384.

\bibitem{wu2024ptv3}
X.~Wu, L.~Jiang, P.-S. Wang \emph{et~al.}, ``Point transformer {V3}: Simpler,
  faster, stronger,'' in \emph{Proc. IEEE/CVF Conf. Comput. Vis. Pattern
  Recognit. (CVPR)}, 2024, pp. 4840--4851.

\bibitem{yuelitept2026}
Y.~Yue, D.~Robert, J.~Wang \emph{et~al.}, ``{LitePT}: Lighter yet stronger
  point transformer,'' in \emph{Proc. IEEE/CVF Conf. Comput. Vis. Pattern
  Recognit. (CVPR)}, 2026, pp. 24\,173--24\,184.

\bibitem{majumdar2023we}
A.~Majumdar, K.~Yadav, S.~Arnaud \emph{et~al.}, ``Where are we in the search
  for an artificial visual cortex for embodied intelligence?'' in \emph{Proc.
  Adv. Neural Inf. Process. Syst. (NeurIPS)}, vol.~36, 2023, pp. 655--677.

\bibitem{ma2022vip}
Y.~J. Ma, S.~Sodhani, D.~Jayaraman, O.~Bastani, V.~Kumar, and A.~Zhang,
  ``{VIP}: Towards universal visual reward and representation via
  value-implicit pre-training,'' in \emph{Proc. Int. Conf. Learn. Represent.
  (ICLR)}, 2023.

\bibitem{radosavovic2023real}
I.~Radosavovic, T.~Xiao, S.~James, P.~Abbeel, J.~Malik, and T.~Darrell,
  ``Real-world robot learning with masked visual pre-training,'' in \emph{Proc.
  Conf. Robot Learn. (CoRL)}, ser. PMLR, vol. 205, 2023, pp. 416--426.

\bibitem{seo2023multi}
Y.~Seo, J.~Kim, S.~James, K.~Lee, J.~Shin, and P.~Abbeel, ``Multi-view masked
  world models for visual robotic manipulation,'' in \emph{Proc. Int. Conf.
  Mach. Learn. (ICML)}, ser. PMLR, vol. 202, 2023, pp. 30\,613--30\,632.

\bibitem{qian20243d}
S.~Qian, K.~Mo, V.~Blukis, D.~F. Fouhey, D.~Fox, and A.~Goyal, ``{3D-MVP}: {3D}
  multiview pretraining for manipulation,'' in \emph{Proc. IEEE/CVF Conf.
  Comput. Vis. Pattern Recognit. (CVPR)}, 2025, pp. 22\,530--22\,539.

\bibitem{hou20254d}
C.~Hou, Y.~Ze, Y.~Fu \emph{et~al.}, ``{4D} visual pre-training for robot
  learning,'' in \emph{Proc. IEEE/CVF Int. Conf. Comput. Vis. (ICCV)}, 2025,
  pp. 8451--8461.

\bibitem{di2024dinobot}
N.~Di~Palo and E.~Johns, ``{DINOBot}: Robot manipulation via retrieval and
  alignment with vision foundation models,'' in \emph{Proc. IEEE Int. Conf.
  Robot. Autom. (ICRA)}, 2024, pp. 2798--2805.

\bibitem{yurchyk2025large}
H.~Yurchyk, W.-D. Chang, G.~Dudek, and D.~Meger, ``Large pre-trained models for
  bimanual manipulation in {3D},'' in \emph{Proc. IEEE-RAS Int. Conf. Humanoid
  Robots (Humanoids)}, 2025, pp. 1195--1202.

\bibitem{zhou2024dino}
G.~Zhou, H.~Pan, Y.~LeCun, and L.~Pinto, ``{DINO-WM}: World models on
  pre-trained visual features enable zero-shot planning,'' in \emph{Proc. Int.
  Conf. Mach. Learn. (ICML)}, ser. PMLR, vol. 267, 2025, pp. 79\,115--79\,135.

\bibitem{loquercio2021learning}
A.~Loquercio, E.~Kaufmann, R.~Ranftl, M.~M{\"u}ller, V.~Koltun, and
  D.~Scaramuzza, ``Learning high-speed flight in the wild,'' \emph{Sci.
  Robot.}, vol.~6, no.~59, p. eabg5810, 2021.

\bibitem{yu2025depth}
H.~Yu, C.~De~Wagter, and G.~C. H.~E. de~Croon, ``Depth transfer: Learning to
  see like a simulator for real-world drone navigation,'' \emph{IEEE Robot.
  Autom. Lett.}, vol.~10, no.~11, pp. 11\,848--11\,855, 2025.

\bibitem{zhang2024resilient}
C.~Zhang, J.~Jin, J.~Frey \emph{et~al.}, ``Resilient legged local navigation:
  Learning to traverse with compromised perception end-to-end,'' in \emph{Proc.
  IEEE Int. Conf. Robot. Autom. (ICRA)}, 2024, pp. 34--41.

\bibitem{he2025attention}
J.~He, C.~Zhang, F.~Jenelten, R.~Grandia, M.~B{\"a}cher, and M.~Hutter,
  ``Attention-based map encoding for learning generalized legged locomotion,''
  \emph{Sci. Robot.}, vol.~10, no. 105, p. eadv3604, 2025.

\bibitem{rudin2025parkour}
N.~Rudin, J.~He, J.~Aurand, and M.~Hutter, ``Parkour in the wild: Learning a
  general and extensible agile locomotion policy using multi-expert
  distillation and {RL} fine-tuning,'' \emph{Int. J. Robot. Res.}, 2026.

\bibitem{luo2024pie}
S.~Luo, S.~Li, R.~Yu, Z.~Wang, J.~Wu, and Q.~Zhu, ``{PIE}: Parkour with
  implicit-explicit learning framework for legged robots,'' \emph{IEEE Robot.
  Autom. Lett.}, vol.~9, no.~11, pp. 9986--9993, 2024.

\bibitem{kareer2022vinl}
S.~Kareer, N.~Yokoyama, D.~Batra, S.~Ha, and J.~Truong, ``{ViNL}: Visual
  navigation and locomotion over obstacles,'' in \emph{Proc. IEEE Int. Conf.
  Robot. Autom. (ICRA)}, 2023, pp. 2018--2024.

\bibitem{zhuang2024humanoid}
Z.~Zhuang, S.~Yao, and H.~Zhao, ``Humanoid parkour learning,'' in \emph{Proc.
  Conf. Robot Learn. (CoRL)}, ser. PMLR, vol. 270, 2025, pp. 1975--1991.

\bibitem{sun2025dpl}
J.~Sun, G.~Han, P.~Sun \emph{et~al.}, ``{DPL}: Depth-only perceptive humanoid
  locomotion via realistic depth synthesis and cross-attention terrain
  reconstruction,'' \emph{IEEE Robot. Autom. Lett.}, vol.~11, no.~8, pp.
  9407--9414, 2026.

\bibitem{savva2019habitat}
M.~Savva, A.~Kadian, O.~Maksymets \emph{et~al.}, ``Habitat: A platform for
  embodied {AI} research,'' in \emph{Proc. IEEE/CVF Int. Conf. Comput. Vis.
  (ICCV)}, 2019, pp. 9338--9346.

\bibitem{wang2020tartanair}
W.~Wang, D.~Zhu, X.~Wang \emph{et~al.}, ``{TartanAir}: A dataset to push the
  limits of visual {SLAM},'' in \emph{Proc. IEEE/RSJ Int. Conf. Intell. Robots
  Syst. (IROS)}, 2020, pp. 4909--4916.

\bibitem{james2022coarse}
S.~James, K.~Wada, T.~Laidlow, and A.~J. Davison, ``Coarse-to-fine
  {Q}-attention: Efficient learning for visual robotic manipulation via
  discretisation,'' in \emph{Proc. IEEE/CVF Conf. Comput. Vis. Pattern
  Recognit. (CVPR)}, 2022, pp. 13\,729--13\,738.

\bibitem{chen2023visualdex}
T.~Chen, M.~Tippur, S.~Wu, V.~Kumar, E.~Adelson, and P.~Agrawal, ``Visual
  dexterity: In-hand reorientation of novel and complex object shapes,''
  \emph{Sci. Robot.}, vol.~8, no.~84, p. eadc9244, 2023.

\bibitem{pitz2024learning}
J.~Pitz, L.~R{\"o}stel, L.~Sievers, D.~Burschka, and B.~B{\"a}uml, ``Learning a
  shape-conditioned agent for purely tactile in-hand manipulation of various
  objects,'' in \emph{Proc. IEEE/RSJ Int. Conf. Intell. Robots Syst. (IROS)},
  2024, pp. 13\,112--13\,119.

\bibitem{zeng2018learning}
A.~Zeng, S.~Song, S.~Welker, J.~Lee, A.~Rodriguez, and T.~Funkhouser,
  ``Learning synergies between pushing and grasping with self-supervised deep
  reinforcement learning,'' in \emph{Proc. IEEE/RSJ Int. Conf. Intell. Robots
  Syst. (IROS)}, 2018, pp. 4238--4245.

\bibitem{kalashnikov2018scalable}
D.~Kalashnikov, A.~Irpan, P.~Pastor \emph{et~al.}, ``Scalable deep
  reinforcement learning for vision-based robotic manipulation,'' in
  \emph{Proc. Conf. Robot Learn. (CoRL)}, ser. PMLR, vol.~87, 2018, pp.
  651--673.

\bibitem{lin2025sim}
T.~Lin, K.~Sachdev, L.~Fan, J.~Malik, and Y.~Zhu, ``Sim-to-real reinforcement
  learning for vision-based dexterous manipulation on humanoids,'' in
  \emph{Proc. Conf. Robot Learn. (CoRL)}, ser. PMLR, vol. 305, 2025, pp.
  4926--4940.

\bibitem{singh2024dextrah}
R.~Singh, A.~Allshire, A.~Handa, N.~Ratliff, and K.~Van~Wyk, ``{DextrAH-RGB}:
  Visuomotor policies to grasp anything with dexterous hands,'' in \emph{Proc.
  IEEE Int. Conf. Robot. Autom. (ICRA)}, 2026, pp. 5086--5093.

\bibitem{he2025viral}
T.~He, Z.~Wang, H.~Xue \emph{et~al.}, ``{VIRAL}: Visual sim-to-real at scale
  for humanoid loco-manipulation,'' in \emph{Proc. IEEE/CVF Conf. Comput. Vis.
  Pattern Recognit. (CVPR)}, 2026, pp. 13\,430--13\,441.

\bibitem{mittal2025isaac}
M.~Mittal, P.~Roth, J.~Tigue \emph{et~al.}, ``{Isaac Lab}: A {GPU}-accelerated
  simulation framework for multi-modal robot learning,'' \emph{arXiv preprint
  arXiv:2511.04831}, 2025.

\bibitem{mittal2023orbit}
M.~Mittal, C.~Yu, Q.~Yu \emph{et~al.}, ``{Orbit}: A unified simulation
  framework for interactive robot learning environments,'' \emph{IEEE Robot.
  Autom. Lett.}, vol.~8, no.~6, pp. 3740--3747, 2023.

\bibitem{tao2024maniskill3}
S.~Tao, F.~Xiang, A.~Shukla \emph{et~al.}, ``Demonstrating {GPU} parallelized
  robot simulation and rendering for generalizable embodied {AI} with
  {ManiSkill3},'' in \emph{Proc. Robot.: Sci. Syst. (RSS)}, 2025.

\bibitem{sablayrolles2018spreading}
A.~Sablayrolles, M.~Douze, C.~Schmid, and H.~J{\'e}gou, ``Spreading vectors for
  similarity search,'' in \emph{Proc. Int. Conf. Learn. Represent. (ICLR)},
  2019.

\bibitem{deng2009imagenet}
J.~Deng, W.~Dong, R.~Socher, L.-J. Li, K.~Li, and L.~Fei-Fei, ``{ImageNet}: A
  large-scale hierarchical image database,'' in \emph{Proc. IEEE Conf. Comput.
  Vis. Pattern Recognit. (CVPR)}, 2009, pp. 248--255.

\bibitem{yang2024depth}
L.~Yang, B.~Kang, Z.~Huang \emph{et~al.}, ``{Depth Anything V2},'' in
  \emph{Proc. Adv. Neural Inf. Process. Syst. (NeurIPS)}, vol.~37, 2024, pp.
  21\,875--21\,911.

\bibitem{straub2019replica}
J.~Straub, T.~Whelan, L.~Ma \emph{et~al.}, ``The {Replica} dataset: A digital
  replica of indoor spaces,'' \emph{arXiv preprint arXiv:1906.05797}, 2019.

\bibitem{roberts2021hypersim}
M.~Roberts, J.~Ramapuram, A.~Ranjan \emph{et~al.}, ``{Hypersim}: A
  photorealistic synthetic dataset for holistic indoor scene understanding,''
  in \emph{Proc. IEEE/CVF Int. Conf. Comput. Vis. (ICCV)}, 2021, pp.
  10\,892--10\,902.

\bibitem{gilles2023metagraspnetv2}
M.~Gilles, Y.~Chen, E.~Z. Zeng \emph{et~al.}, ``{MetaGraspNetV2}: All-in-one
  dataset enabling fast and reliable robotic bin picking via object
  relationship reasoning and dexterous grasping,'' \emph{IEEE Trans. Autom.
  Sci. Eng.}, vol.~21, no.~3, pp. 2302--2320, 2024.

\bibitem{patel2025tartanground}
M.~Patel, F.~Yang, Y.~Qiu \emph{et~al.}, ``{TartanGround}: A large-scale
  dataset for ground robot perception and navigation,'' in \emph{Proc. IEEE/RSJ
  Int. Conf. Intell. Robots Syst. (IROS)}, 2025, pp. 20\,524--20\,531.

\bibitem{sun2022shift}
T.~Sun, M.~Segu, J.~Postels \emph{et~al.}, ``{SHIFT}: A synthetic driving
  dataset for continuous multi-task domain adaptation,'' in \emph{Proc.
  IEEE/CVF Conf. Comput. Vis. Pattern Recognit. (CVPR)}, 2022, pp.
  21\,339--21\,350.

\bibitem{zamir2018taskonomy}
A.~R. Zamir, A.~Sax, W.~Shen, L.~J. Guibas, J.~Malik, and S.~Savarese,
  ``{Taskonomy}: Disentangling task transfer learning,'' in \emph{Proc.
  IEEE/CVF Conf. Comput. Vis. Pattern Recognit. (CVPR)}, 2018, pp. 3712--3722.

\bibitem{ramakrishnan2021habitat}
S.~K. Ramakrishnan, A.~Gokaslan, E.~Wijmans \emph{et~al.},
  ``{Habitat-Matterport 3D Dataset (HM3D)}: 1000 large-scale {3D} environments
  for embodied {AI},'' in \emph{Proc. Adv. Neural Inf. Process. Syst. Datasets
  Benchmarks Track}, vol.~1, 2021.

\bibitem{baruch2021arkitscenes}
G.~Baruch, Z.~Chen, A.~Dehghan \emph{et~al.}, ``{ARKitScenes}: A diverse
  real-world dataset for {3D} indoor scene understanding using mobile {RGB-D}
  data,'' in \emph{Proc. Adv. Neural Inf. Process. Syst. Datasets Benchmarks
  Track}, vol.~1, 2021.

\bibitem{dai2017scannet}
A.~Dai, A.~X. Chang, M.~Savva, M.~Halber, T.~Funkhouser, and M.~Nie{\ss}ner,
  ``{ScanNet}: Richly-annotated {3D} reconstructions of indoor scenes,'' in
  \emph{Proc. IEEE Conf. Comput. Vis. Pattern Recognit. (CVPR)}, 2017, pp.
  2432--2443.

\bibitem{fang2020graspnet}
H.-S. Fang, C.~Wang, M.~Gou, and C.~Lu, ``{GraspNet-1Billion}: A large-scale
  benchmark for general object grasping,'' in \emph{Proc. IEEE/CVF Conf.
  Comput. Vis. Pattern Recognit. (CVPR)}, 2020, pp. 11\,441--11\,450.

\bibitem{xiao2013sun3d}
J.~Xiao, A.~Owens, and A.~Torralba, ``{SUN3D}: A database of big spaces
  reconstructed using {SfM} and object labels,'' in \emph{Proc. IEEE Int. Conf.
  Comput. Vis. (ICCV)}, 2013, pp. 1625--1632.

\bibitem{khazatsky2024droid}
A.~Khazatsky, K.~Pertsch, S.~Nair \emph{et~al.}, ``{DROID}: A large-scale
  in-the-wild robot manipulation dataset,'' in \emph{Proc. Robot.: Sci. Syst.
  (RSS)}, 2024.

\bibitem{frey2025boxi}
J.~Frey, T.~Tuna, L.~F.~T. Fu \emph{et~al.}, ``{Boxi}: Design decisions in the
  context of algorithmic performance for robotics,'' in \emph{Proc. Robot.:
  Sci. Syst. (RSS)}, 2025.

\bibitem{cho2021diml}
J.~Cho, D.~Min, Y.~Kim, and K.~Sohn, ``{DIML/CVL} {RGB-D} dataset: {2M} {RGB-D}
  images of natural indoor and outdoor scenes,'' \emph{arXiv preprint
  arXiv:2110.11590}, 2021.

\bibitem{yang2019drivingstereo}
G.~Yang, X.~Song, C.~Huang, Z.~Deng, J.~Shi, and B.~Zhou, ``{DrivingStereo}: A
  large-scale dataset for stereo matching in autonomous driving scenarios,'' in
  \emph{Proc. IEEE/CVF Conf. Comput. Vis. Pattern Recognit. (CVPR)}, 2019, pp.
  899--908.

\bibitem{fu2018deep}
H.~Fu, M.~Gong, C.~Wang, K.~Batmanghelich, and D.~Tao, ``Deep ordinal
  regression network for monocular depth estimation,'' in \emph{Proc. IEEE/CVF
  Conf. Comput. Vis. Pattern Recognit. (CVPR)}, 2018, pp. 2002--2011.

\bibitem{hinton2015distilling}
G.~Hinton, O.~Vinyals, and J.~Dean, ``Distilling the knowledge in a neural
  network,'' \emph{arXiv preprint arXiv:1503.02531}, 2015.

\bibitem{schulman2017proximal}
J.~Schulman, F.~Wolski, P.~Dhariwal, A.~Radford, and O.~Klimov, ``Proximal
  policy optimization algorithms,'' \emph{arXiv preprint arXiv:1707.06347},
  2017.

\bibitem{he2016deep}
K.~He, X.~Zhang, S.~Ren, and J.~Sun, ``Deep residual learning for image
  recognition,'' in \emph{Proc. IEEE Conf. Comput. Vis. Pattern Recognit.
  (CVPR)}, 2016, pp. 770--778.

\bibitem{radosavovic2020designing}
I.~Radosavovic, R.~P. Kosaraju, R.~Girshick, K.~He, and P.~Doll{\'a}r,
  ``Designing network design spaces,'' in \emph{Proc. IEEE/CVF Conf. Comput.
  Vis. Pattern Recognit. (CVPR)}, 2020, pp. 10\,425--10\,433.

\bibitem{tan2019efficientnet}
M.~Tan and Q.~Le, ``{EfficientNet}: Rethinking model scaling for convolutional
  neural networks,'' in \emph{Proc. Int. Conf. Mach. Learn. (ICML)}, ser. PMLR,
  vol.~97, 2019, pp. 6105--6114.

\bibitem{tan2020efficientdet}
M.~Tan, R.~Pang, and Q.~V. Le, ``{EfficientDet}: Scalable and efficient object
  detection,'' in \emph{Proc. IEEE/CVF Conf. Comput. Vis. Pattern Recognit.
  (CVPR)}, 2020, pp. 10\,778--10\,787.

\bibitem{song2015sun}
S.~Song, S.~P. Lichtenberg, and J.~Xiao, ``{SUN RGB-D}: A {RGB-D} scene
  understanding benchmark suite,'' in \emph{Proc. IEEE Conf. Comput. Vis.
  Pattern Recognit. (CVPR)}, 2015, pp. 567--576.

\bibitem{neigel2021offsed}
P.~Neigel, J.~Rambach, and D.~Stricker, ``{OFFSED}: Off-road semantic
  segmentation dataset,'' in \emph{Proc. Int. Joint Conf. Comput. Vis. Imag.
  Comput. Graph. Theory Appl. (VISAPP)}.\hskip 1em plus 0.5em minus 0.4em\relax
  SciTePress, 2021, pp. 552--557.

\bibitem{vogel2024robust}
D.~Vogel, R.~Baines, J.~Church, J.~Lotzer, K.~Werner, and M.~Hutter, ``Robust
  ladder climbing with a quadrupedal robot,'' in \emph{Proc. IEEE/RSJ Int.
  Conf. Intell. Robots Syst. (IROS)}, 2025, pp. 7239--7244.

\bibitem{handa2014kinect}
A.~Handa, T.~Whelan, J.~McDonald, and A.~J. Davison, ``A benchmark for {RGB-D}
  visual odometry, {3D} reconstruction and {SLAM},'' in \emph{Proc. IEEE Int.
  Conf. Robot. Autom. (ICRA)}, 2014, pp. 1524--1531.

\bibitem{makoviychuk2021isaac}
V.~Makoviychuk, L.~Wawrzyniak, Y.~Guo \emph{et~al.}, ``{Isaac Gym}: High
  performance {GPU} based physics simulation for robot learning,'' in
  \emph{Proc. Adv. Neural Inf. Process. Syst. Datasets Benchmarks Track},
  vol.~1, 2021.

\bibitem{darcet2023vision}
T.~Darcet, M.~Oquab, J.~Mairal, and P.~Bojanowski, ``Vision transformers need
  registers,'' in \emph{Proc. Int. Conf. Learn. Represent. (ICLR)}, 2024.

\end{thebibliography}
